%% file: arxiv.tex
\documentclass{article}
\usepackage[preprint]{neurips_2026}

\usepackage[utf8]{inputenc}
\usepackage[T1]{fontenc}
\usepackage[table]{xcolor}
\usepackage{hyperref}
\usepackage{url}
\usepackage{microtype}
\usepackage{graphicx}
\graphicspath{{../}{./}}
\usepackage{subcaption}
\usepackage{booktabs}
\usepackage{amsmath}
\usepackage{amssymb}
\usepackage{mathtools}
\usepackage{amsthm}
\usepackage[capitalize,noabbrev]{cleveref}
\usepackage{xspace}
\usepackage{tcolorbox}
\usepackage{adjustbox}
\usepackage{array}
\usepackage{algorithm}
\usepackage{algorithmic}
\usepackage{placeins}

\newcommand{\methodfigwidth}{0.95}   
\newcommand{\attnfigwidth}{0.71}     
\newcommand{\cropsfigwidth}{0.69}    
\newcommand{\compfigwidth}{0.69}     

\theoremstyle{plain}

\theoremstyle{definition}

\theoremstyle{remark}

\newcommand{\methodName}{COMiT\xspace}
\newcommand{\methodNameExtended}{COMmunication inspired Tokenization}
\newcommand{\R}{\mathbb{R}\xspace}

\newcommand{\E}{\mathbb{E}\xspace}

\definecolor{second}{RGB}{235, 247, 255}
\definecolor{best}{RGB}{210, 235, 255}
\newcommand{\best}[1]{\cellcolor{best}\textbf{#1}}
\newcommand{\second}[1]{\cellcolor{second}\textbf{#1}}

\usepackage[normalem]{ulem}
\newif\ifrevisionmarkup
\revisionmarkuptrue
\definecolor{revaddcolor}{RGB}{0, 90, 200}
\definecolor{revdelcolor}{RGB}{190, 30, 30}

\makeatletter
\newcommand{\gobblesectionheading}{\def\section{\@ifstar{\@gobble}{\@dblarg\@gobble}}}
\makeatother

\title{COMiT: Learning Structured Visual Tokens through Sequential Communication}

\author{%
  \begin{tabular}{@{}c@{}}
  Aram Davtyan$^{1,*,\dagger}$ \quad Yusuf Sahin$^{1,*}$ \quad Yasaman Haghighi$^{2}$ \quad Sebastian Stapf$^{1}$ \\
  Pablo Acuaviva$^{1}$ \quad Alexandre Alahi$^{2}$ \quad Paolo Favaro$^{1}$ \\[6pt]
  \normalfont $^{1}$Computer Vision Group, University of Bern, Switzerland \quad $^{2}$VITA Lab, EPFL, Switzerland
  \end{tabular}
}

\begin{document}

\maketitle

\begingroup
\renewcommand{\thefootnote}{$*$}\footnotetext{Equal contribution.}
\renewcommand{\thefootnote}{$\dagger$}\footnotetext{Corresponding author: \texttt{aram.davtyan@unibe.ch}}
\endgroup

\begin{center}
  \vspace{-1.5em}
  {\small Website: \url{https://araachie.github.io/comit}}
\end{center}

\input{sections_arxiv/main_paper/abstract}
\input{sections_arxiv/main_paper/introduction}

\input{sections_arxiv/main_paper/prior_work}
\input{sections_arxiv/main_paper/method}
\input{sections_arxiv/main_paper/experiments}

\input{sections_arxiv/main_paper/conclusion}
\clearpage
\begin{ack}
\begingroup
\gobblesectionheading
\input{sections_arxiv/main_paper/acknowledgements}
\endgroup
\end{ack}

{\small
\bibliographystyle{comit_references}
\bibliography{common/references}

\newpage
\appendix

\input{sections_arxiv/appendix/additional_implementation_details}
\input{sections_arxiv/appendix/additional_evaluation_details}
\input{sections_arxiv/appendix/attn_maps}

\input{sections_arxiv/appendix/additional_ablations}
\input{sections_arxiv/appendix/extended_related_work}
\input{sections_arxiv/appendix/more_visual_results}
\input{sections_arxiv/appendix/future_work}
\input{sections_arxiv/appendix/impact_statement}

\FloatBarrier


\end{document}

%% file: sections_arxiv/main_paper/abstract.tex
\begin{abstract}
Discrete image tokenizers provide a sequential interface for vision and multimodal models, but are typically optimized for reconstruction or compression and therefore tend to encode local appearance rather than object-level structure. We introduce \methodName, a communication-inspired framework for learning structured discrete visual representations. \methodName constructs a fixed-length latent message through sequential visual observations: at each step, a transformer processes a localized image crop and updates, refines, and reorganizes the existing token sequence. After several iterations, the resulting message conditions a flow-matching decoder that reconstructs the complete image. The encoder and decoder are implemented within a single transformer and trained end-to-end using flow-matching reconstruction and semantic representation-alignment objectives. \methodName substantially improves compositional generalization and relational reasoning over prior methods. Our experiments show that, while semantic alignment helps ground the representation, attentive sequential tokenization is critical for inducing more interpretable, object-centric token structures.
\end{abstract}

%% file: sections_arxiv/main_paper/introduction.tex
\section{Introduction}

Modern multimodal systems increasingly process vision through the lens of sequence modeling~\citep{sun2024autoregressive, chen2025janus}: images are converted into discrete token sequences that can be consumed by transformer-based architectures. This token-based interface enables scalable training, efficient compression, and unified reasoning over visual and textual inputs. Consequently, learning effective image tokenizers has emerged as a central problem in visual representation learning.

Conventional discrete encoders~\citep{van2017neural, esser2021taming} typically represent images as two-dimensional grids of tokens and are trained primarily for reconstruction under a compression constraint. As a result, the learned tokens often capture local texture and patch statistics rather than object-level semantic structure, limiting their interpretability and usefulness for downstream image understanding tasks.

Recent progress in image tokenization has renewed interest in \emph{one-dimensional} discrete bottlenecks, which better match the sequential format expected by transformer-based models and can improve the semantic organization of token sequences~\citep{yu2024image, wang2025selftok, bachmann2025flextok, duggal2024adaptive}. However, despite this syntactic compatibility, most existing approaches remain primarily optimized for compression trade-offs. As a result, semantic information is often entangled and poorly localized across tokens, limiting performance on downstream tasks that require compositional, object-centric structure.

In this work, we shift the focus from compression trade-offs to the \emph{semantic organization} of visual token sequences. While semantic information can be encouraged through alignment with pretrained visual representations, we argue that such supervision alone is insufficient to induce structured and interpretable tokens. Instead, structured visual tokenization requires coupling semantic learning objectives with an encoding procedure that explicitly encourages compositional organization.

Our approach is inspired by how humans communicate about visual scenes. When describing a scene containing multiple objects, a speaker typically attends to one region at a time, sequentially incorporating salient information into the message. This incremental process allows the listener to integrate each observation into a coherent mental model of the scene while gradually reducing uncertainty~\citep{hassabis2007deconstructing}. Under limited communication bandwidth, descriptions naturally prioritize high-level entities and their relations over fine-grained details, yielding a certain coarse-to-fine hierarchy in the space of possible descriptions. Notably, while so far we have considered the speaker and the listener to be two separate roles, humans are able to simultaneously perform both tasks. For instance, one could imagine the setting where the speaker is trying to memorize the scene for a certain period of time and then is asked to recall it.

Motivated by this perspective, we study how a communication-inspired encoding process can shape the structure of learned visual token sequences. In particular, we adopt two key design principles:

\begin{tcolorbox}[colback=best, colframe=second, width=\linewidth, boxrule=0pt, arc=0pt]
\begin{itemize}
    \item \textbf{Attentive and sequential tokenization.} The encoder processes the image as a sequence of localized observations, attending to different regions at each step and incrementally integrating information into a discrete latent message.
    \item \textbf{Homogeneous communication.} In contrast to traditional autoencoders that use separate encoder and decoder networks, we adopt a unified design in which the same network acts as both ``speaker'' and ``listener'', mirroring the symmetry in human communication.
\end{itemize}
\end{tcolorbox}

Building on these principles, we propose \emph{\methodNameExtended} (\methodName), which formulates image encoding as an iterative communication-and-reconstruction game. At each step, the model observes a new image crop and updates its discrete latent message. After several steps, the final message serves as a compact representation of the scene, from which the same network reconstructs the full image. The whole pipeline is trained as a generative model within the flow-matching framework, so that reconstruction acts as a learning signal rather than the end goal: we position \methodName as an understanding-oriented tokenizer with reconstruction capability rather than as a state-of-the-art reconstruction codec. To encourage semantic grounding, we incorporate a semantic representation alignment objective that distills high-level features from a frozen self-supervised vision model. Crucially, while alignment provides semantic signal, the attentive sequential tokenization determines how this information is distributed and localized across tokens.

To evaluate these properties, we introduce a suite of benchmarks that probe not only semantic content, but also compositional generalization and relational reasoning. Across these evaluations, \methodName consistently outperforms existing one-dimensional discrete image encoders by a substantial margin. Ablations further reveal complementary effects: semantic alignment improves token meaning, while communication-inspired sequential encoding induces more interpretable, object-centric tokens.

%% file: sections_arxiv/main_paper/prior_work.tex
\section{Related Work}\label{sec:related_work}


\textbf{Attentive Encoding.} The idea of aggregating information from partial observations of a scene has a long history. One early example is the Recurrent Attention Model (RAM) proposed by \citet{mnih2014recurrent}, where a model iteratively selects regions of an image to attend to and accumulates a latent state to solve classification. In that work, the primary motivation was computational efficiency on large images, rather than representation learning or reconstruction. In other early approaches \citep{eslami2016attend, eslami2018nsr} the reconstruction serves as the main training signal. However, often the latent state in such models explicitly encodes inductive biases, such as object-level bounding box coordinates~\citep{eslami2016attend}, rather than abstract concepts. Moreover, the early methods were only tested on rather toy data, and, to the best of our knowledge, there is no evidence on scaling these to real images.

\textbf{Image Tokenization.} Early works such as VQ-VAE~\citep{van2017neural, razavi2019generating} introduced discrete visual codebooks learned via vector quantization. Subsequent approaches, including MaskGIT~\citep{chang2022maskgit} and VQ-GAN~\citep{esser2021taming}, improved generative fidelity through masked prediction and adversarial training. More recently, several works have explored \emph{one-dimensional} tokenization schemes, with \citet{yu2024image} pioneering this direction. Many of these methods, however, continue to treat tokenization primarily as a compression and reconstruction objective, producing tokens in a single encoding pass without explicit mechanisms for sequential refinement. Other works have investigated how token ordering can reflect properties of the underlying visual signal, such as semantic hierarchies~\citep{bachmann2025flextok} or frequency structure~\citep{wang2025selftok}. Closest to our approach, \citet{duggal2024adaptive} also introduce an iterative recurrent procedure for refining the latent representation. In their setting, however, the first encoding step is designed to capture most of the image content, while subsequent refinement steps use additionally allocated capacity. Overall, existing tokenizers remain largely optimized for reconstruction-compression trade-offs rather than structure and semantics. Appendix~\ref{app:extended_related_work} expands this discussion with more recent and concurrent tokenizers.

\textbf{Flow Matching and Diffusion Autoencoders.} Flow-matching generative models \citep{lipmanflow, liu2022flow} offer stable and efficient training of continuous data distributions. These frameworks have also been used to train decoders in autoencoders \citep{preechakul2022diffusion, guo2025variational, bachmann2025flextok, chen2025diffusion}. However, to the best of our knowledge, we are the first to leverage this formulation to train both encoding and decoding stages through a unified, differentiable flow objective, integrating representation learning and generation into a single network.

%% file: sections_arxiv/main_paper/method.tex
\section{Communication Inspired Tokenization}\label{sec:method}

\begin{figure}[t]
    \centering
    \includegraphics[width=\methodfigwidth\linewidth]{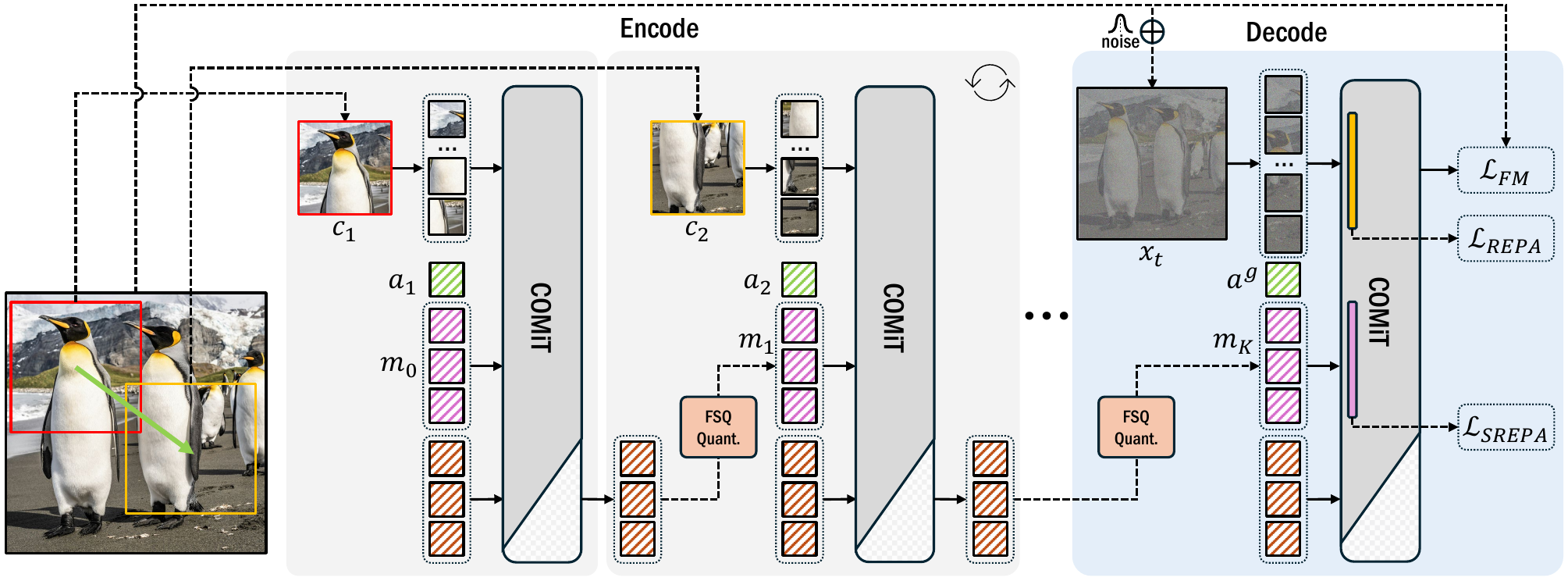}
    \caption{The overall training pipeline of \methodName. A sequence of $K$ random crops is extracted from the input image and iteratively embedded into the latent message $m_K$ that is discretized via FSQ~\cite{mentzerfinite}. The latter is decoded by the same model using the flow matching objective~\cite{lipmanflow}. Additionally, we use REPA~\cite{yu2024representation} to speed up the training and SREPA to inject more semantic priors into the latent message.}
    \label{fig:method}
\end{figure}

In this section we start by describing the high-level training pipeline of \methodName that is illustrated in Figure~\ref{fig:method}. We then talk about some crucial details in Sections~\ref{sec:greedy_use_of_tokens}--\ref{sec:cropping_policies}.

\noindent\textbf{Encoding.} Given an input RGB\footnote{In fact, \methodName works in the latent space of a pretrained 2D VAE, but we omit the VAE in our notation as it does not affect any of the training steps.} image $x \in \R^{3\times H\times W}$, we define a sequence of random crops $\{c_k\}_{k=1}^K, \; c_k \in \R^{3 \times h \times w}$ and their corresponding locations $\{l_k\}_{k = 1}^K, \; l_k \in [-1, 1]^2$. Here $l_k$ is the absolute location of the $c_k$ crop's center in the image $x$ with each component normalized to $[-1, 1]$. From these, we derive the sequence of offsets $\{a_k\}_{k = 1}^K$, where $a_1 = (0, 0)$ and $a_k = l_k - l_{k - 1}, k \geq 2$. The offsets can be interpreted as actions the agent takes to scan the scene. We opt for relative offsets instead of global grounding to prevent parts of the latent message to specialize on certain constant regions. 

In the beginning, the latent message $m_0 \in \mathbb{R}^{L\times d}$ is initialized with the same $L$ $d$-dimensional vectors from a dictionary of $N$ unique tokens. We refer to $L$ as the message length and $N$ as the vocabulary size. After that, at each step, the model $f_\theta$ takes as input the current latent message $m_{k-1}$ and updates it with the information from the current crop:
\begin{align}
    m_k = f^m_\theta(c_k, t_k, a_k, m_{k-1}).
\end{align}
Here, the superscript $m$ in $f^m_\theta$ denotes the readout part of the model's output that corresponds to the updated message. This is needed to support different output modalities that we introduce below. The second argument in $f_\theta^m$ corresponds to the denoising timestamp, which is $t_k = 1$ for the clean crops. This is needed to distinguish the encoding from the decoding regime, which we also introduce below.

The predicted message $m_k$ is then quantized via FSQ~\cite{mentzerfinite} to project the tokens to the vocabulary and fed back to the model along with the newly observed crop, forming a recurrent loop. After $K$ steps, the final message $m_K$ is obtained.

\noindent\textbf{Decoding.} The message derived through the above procedure of recurrent crop aggregation serves as conditioning for flow-based decoding of the full image. However, in contrast to the conventional autoencoder architectures, \methodName uses the same model for both encoding and decoding. That is, we start by creating a noisy version of the image $x_t = t x + (1 - t)\varepsilon$ with $t \sim U[0, 1]$ -- random interpolation time and $\varepsilon \sim \mathcal{N}(0, I)$ -- random noise instance. Then, the noisy image $x_t$ is fed to the network along with the final message $m_K$ and the offset relative to the last crop $a^g = -l_K$. The network is then supervised to predict the velocity of the marginal flow at $x_t$ via the standard conditional flow-matching loss~\citep{lipmanflow}:
\begin{align}\label{eq:fm}
    \mathcal{L}_{\rm FM} = \underset{\substack{t \sim U[0, 1]\\ \varepsilon \sim \mathcal{N}(0, I)}}{\E} \left\|f^v_\theta(x_t, t, a^g, m_{K}) - (x - \varepsilon)\right\|^2_2. 
\end{align}

Here, as before, the superscript $v$ in $f^v_\theta$ denotes the readout part of the model's output that corresponds to the estimated velocity.


\noindent\textbf{Inference.} Given a latent message $\hat m$, in order to visually probe its content, one may decode the message through numerically integrating the flow ODE~\cite{lipmanflow}. That is, starting at a noise sample $x_0 \sim \mathcal{N}(0, I)$, one obtains the clean image $x_1$ via iteratively denoising the current estimate:
\begin{align}
    x_{t + h} = x_t + h f_\theta^v(x_t, t, a^g, \hat m),
\end{align}
where $h$ is the step size connected to the number of function evaluations (NFE) via $h = 1 / \rm{NFE}$.

\subsection{Greedy Use of Tokens}
\label{sec:greedy_use_of_tokens}
As noted in the introduction, as the amount of information embedded in the latent message increases (in our case, the number of crops), it is natural to expect the model to discard irrelevant secondary details and retain only the most essential features for reconstruction. This process would naturally induce a hierarchical structure in the feature space. However, if during training the number of crops the model aggregates were fixed, the network would instead learn to pre-allocate parts of the latent message for future crops. This would effectively enforce a fixed capacity per crop, with each information chunk occupying a predetermined location in the message. To avoid this behavior, we randomize the number of crops the model processes during training. As a result, the model never knows whether additional crops will follow the current one and is therefore encouraged to use the available tokens greedily. This corresponds to the setting with fixed capacity and varying amount of information, which is precisely the regime we were aiming for (see Appendix~\ref{app:crop_randomization} for ablations on the number of crops).

In addition, backpropagating gradients through the entire sequence of message updates would be computationally expensive and memory-intensive. To address this, we apply a stop-gradient operation to all updates except the final one. Combined with the randomized number of crop aggregations, this further promotes greedy token usage while enabling efficient training. In preliminary small-scale experiments, we found that restricting gradient backpropagation in this way has only a moderate impact on performance (see Appendix~\ref{app:heuristic_robustness} for a full-scale ablation).

\subsection{Distilling Semantic Representation} \label{sec:srepa}

As noted in the introduction, to enforce the semantic ordering of the information in the learned messages, we directly distill pretrained SSL features $\psi(x) \in \mathbb{R}^{s}$(namely, the \texttt{[CLS]} of DINOv2~\cite{oquab2023dinov2}) into \methodName's intermediate representations that correspond to the message tokens. We call this procedure semantic representation alignment, or SREPA. More precisely, during the decoding stage, the intermediate representations in the $j$-th layer of the network $f_\theta^m[j](x_t, t, a^{g}, m_K) \in \mathbb{R}^{L\times r}$ are first projected with a small MLP and then averaged pooled into a single vector $\bar f_\theta^m[j](x_t, t, a^{g}, m_K) \in \mathbb{R}^s$. Then, the following loss is calculated:
\begin{align}\label{eq:srepa}
    \mathcal{L}_{\rm SREPA} =  \exp\left(-{\rm Sim}\left(\psi(x), \bar f_\theta^m[j](x_t, t, a^{g}, m_K)\right)\right),
\end{align}
where $\rm{Sim}(\cdot, \cdot)$ is the cosine similarity.

\subsection{Reconstruction Fidelity}

Although our main focus in this paper is on semantics of the token sequences, we apply several techniques to improve the reconstruction fidelity of \methodName. First, to enable classifier-free guidance (CFG)~\cite{ho2022classifier} at inference, during training we skip the encoding part with probability $p_{\rm CFG} \approx 0.2$, thereby training an unconditional decoder. Besides this, with probability $p_{\rm G} \approx 0.5$ we replace the first crop with the full original image, allowing the model to encode the whole image in a single step. For clarity, we refer to the full image as the \emph{global crop}, and to the smaller crops as \emph{local crops}.

As a result, the training consists of three regimes: in some cases, \methodName first observes the global crop and then refines its latent message with local crops; in others, it observes only local crops; and the third portion of the training is devoted to the decoding of the image from an empty message.

\subsection{Implementation Details} \label{sec:implementation_details}

In practice, \methodName is implemented as a transformer encoder network~\cite{vaswani2017attention}. Following the standard practices of DiT~\cite{peebles2023scalable}, we use AdaLN~\cite{perez2018film} layers to make the network timestamp-conditioned. Notably, different modalities (image, message) use separate projections in the AdaLN layers. All inputs to the network are flattened into a single sequence of tokens. Images are first  encoded with a pretrained VAE (namely, the KL-regularized VAE of Stable Diffusion~\cite{rombach2022high}) to reduce the input size and then patchified. The offset is embedded into a single token via linear projection. We found that the model performs slightly better when, instead of overwriting the previous message, one feeds both the previous message and newly initialized buffer tokens to predict the next message (see also Figure~\ref{fig:method}). This architectural design also allows us to apply causal masks in the attention maps of the buffer tokens, mimicking the causal nature of composing sentences (illustrated in Figure~\ref{fig:method} as the tilted line of the gray shaded region in \methodName's main backbone). 
We train \methodName in three model sizes: B (12 layers, hidden size 768), L (24 layers, hidden size 1024) and XL (28 layers, hidden size 1152). All models are trained on ImageNet1k~\cite{deng2009imagenet}. To speed up the convergence, we use REPA~\cite{yu2024representation} that aligns the intermediate image representation in the network with the DINOv2's spatial features. The final loss thus takes the form:
\begin{align}\label{eq:total_loss}
    \mathcal{L} = \mathcal{L}_{\rm FM} + \lambda_{\rm REPA} \mathcal{L}_{\rm REPA} + \lambda_{\rm SREPA}\mathcal{L}_{\rm SREPA}.
\end{align}
In our experiments, we choose $\lambda_{\rm REPA} = \lambda_{\rm SREPA} = 0.5$.

The whole training procedure is summarized in Figure~\ref{fig:method}. Implementation details are in Appendix~\ref{app:implementation_details}.

\subsection{Cropping Policies}\label{sec:cropping_policies}

Because both the crops and their number are randomized during training, \methodName offers flexibility at inference time in how crops are selected when encoding an image. We refer to these choices as \emph{cropping policies}. Random crops thus only serve as a training distribution that supplies diverse scan paths without annotations. The policy used at test time is a free design choice: it can cover the image exhaustively, follow a fixed order, or select regions based on the model's current uncertainty, and in principle it could also express approximations of human-like scan paths. Although many cropping policies are possible, in this paper, we study only a limited subset to illustrate the impact of the main design decisions, leaving a more exhaustive exploration to future work. For simplicity, we resize all images to a resolution of $256 \times 256$ and consider only $96 \times 96$ crops arranged on a $3 \times 3$ grid. We characterize cropping policies by three factors: (i) whether a global crop is included at the beginning of the crop sequence, (ii) the number of crops, and (iii) the order in which the crops are processed. With respect to the latter, we consider random ordering, a fixed raster-scan order, and an adaptive policy. The adaptive policy decodes the current latent message after each newly aggregated crop and selects the next crop to be the one that has the largest mean reconstruction error. The intermediate reconstructions are obtained using a single decoding step, resulting in blurry regions in areas where the model is uncertain about the content (see Figures~\ref{fig:composition} and \ref{fig:global_vs_local}).

%% file: sections_arxiv/main_paper/experiments.tex
\section{Experiments}\label{sec:experiments}
In this section, we evaluate \methodName to showcase its improved semantic encodings and emergent properties. We design the test suite of three representative benchmarks.

\noindent\textbf{Visual recognition (ImageNet100).} To evaluate whether \methodName encodes high-level semantic information useful for visual recognition, we perform classification probing on ImageNet100 (IN100)~\cite{deng2009imagenet}. Instead of standard linear probing, we adopt an attention-based probing strategy using a lightweight two-layer of self-attention operating over the 1D sequence of tokens produced by our model and the baselines. This design allows the probe to flexibly attend to and aggregate information across the entire token sequence, without assuming that class-discriminative features are localized at a fixed position. Attention probing is therefore better suited for the 1D tokenization setting, where semantic information may be distributed across tokens. It also allows comparing tokenizers that emit different numbers of tokens per image. As an alternative, attention-free probe, Appendix~\ref{app:affine_probe} reports a single affine readout over the flattened token sequence.

\noindent\textbf{Compositional generalization (MSCOCO).} We evaluate compositional generalization using a benchmark derived from MSCOCO~\cite{lin2014microsoft}, focusing on images that contain two object categories. The dataset is split into two disjoint partitions such that both splits share the same set of object categories, while specific object co-occurrences (i.e., object pairs) appear in only one split. We train the probing network on the first split and evaluate it on the second, ensuring that the probe must generalize to unseen object compositions. This setup tests whether object-related information is stored in a disentangled manner across the token sequence (e.g., distributed across different subsets of tokens corresponding to individual objects, rather than being collapsed into a single token that encodes the entire scene). Indeed, if the model mixed the information about two objects in a single token, the probe would have a more difficult job to effectively generalize to unseen object pairs.

\noindent\textbf{Inter-object relations (Visual Genome).} We evaluate the ability of models to encode relational semantics using relationship prediction on Visual Genome (VG)~\cite{krishna2017visual}. Given a batch of images paired with candidate subject–object relations, we test whether the probing network can correctly assign each image to its corresponding relation. This benchmark assesses whether inter-object relationships are explicitly represented in the token sequence, beyond individual object semantics, and whether such relational information can be recovered through lightweight probing.

For more details on our testing suite, we refer the reader to the Appendix~\ref{app:test_suite}.

\subsection{Ablations}\label{sec:ablations}

\textbf{Semantic distillation.} We evaluate the impact of SREPA by comparing the full COMiT-B model against an ablated variant in which $\lambda_{\rm SREPA} = 0$ during training. Both models are evaluated under identical settings on IN100. We observe that SREPA has significant contribution to our tokenizer's performance (see Table~\ref{tab:srepa_ablation}). Note that since FSQ uses the same codebook across model variants, all the differences in performance may be attributed to token co-occurrences. 

\begin{table}[t]
    \centering
    \footnotesize
    \begin{minipage}[t]{0.47\linewidth}
    \centering
    \caption{The effect of SREPA during training. Both models use the same cropping policy (single global crop) at test time. Parentheses: gain over the ablation.}
    \label{tab:srepa_ablation}
    \setlength{\tabcolsep}{2pt}
    \begin{tabular}{@{}lccc@{}}
    \toprule
    & & IN100 & CSSD \\
    Model variant & SREPA & top-1$\uparrow$ & mIoU$\uparrow$ \\
    \midrule
    
    COMiT-B            & \checkmark & \best{82.91\,{\scriptsize(+10.65)}} & \best{0.527\,{\scriptsize(+0.020)}} \\
    COMiT-B            & --         & 72.26 & 0.507 \\
    \bottomrule
    \end{tabular}
    \end{minipage}
    \hfill
    \begin{minipage}[t]{0.49\linewidth}
    \centering
    \caption{The effect of using recurrent local crops aggregation during training. At test time, both models use the same cropping policy with a single global crop. Parentheses: gain over the ablation.}
    \label{tab:no_local_ablation}
    \setlength{\tabcolsep}{2pt}
    \begin{tabular}{@{}lccc@{}}
    \toprule
     & Local crops & IN100 & CSSD \\
     Model & at training & top-1$\uparrow$ & mIoU$\uparrow$ \\
    \midrule
     \methodName-B & \checkmark & \best{82.91\,{\scriptsize(+1.97)}} & \best{0.527\,{\scriptsize(+0.189)}} \\
     \methodName-B & -- & 80.94 & 0.338 \\
    \bottomrule
    \end{tabular}
    \end{minipage}
\end{table}

\begin{figure}[t]
    \centering
    \footnotesize
    \newcommand{\curWidth}{\attnfigwidth\linewidth}
    \begin{tabular}{@{}c@{}}
        Input images \\
         \includegraphics[width=\curWidth]{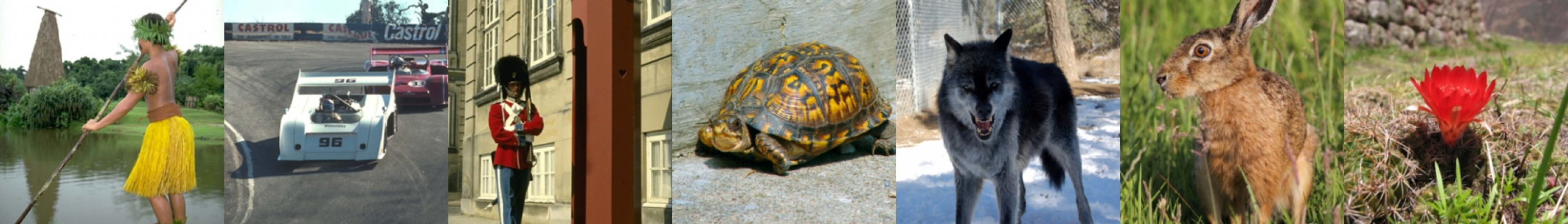} \\
         Ground truth masks \\
         \includegraphics[width=\curWidth]{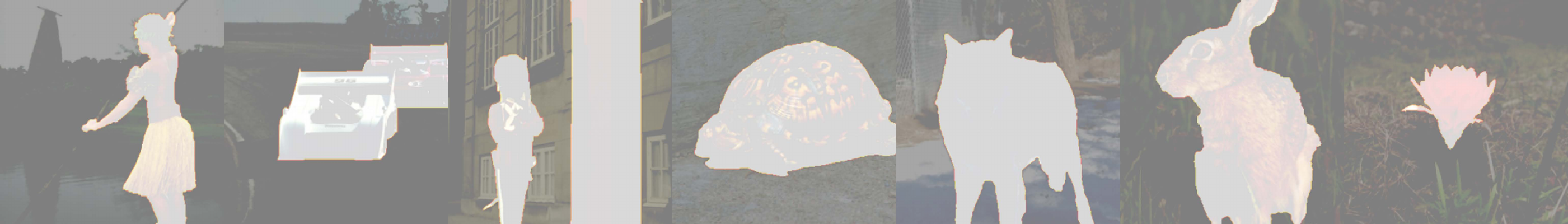} \\
         \methodName-B, ${\rm mIoU} = 0.53$ \\
         \includegraphics[width=\curWidth]{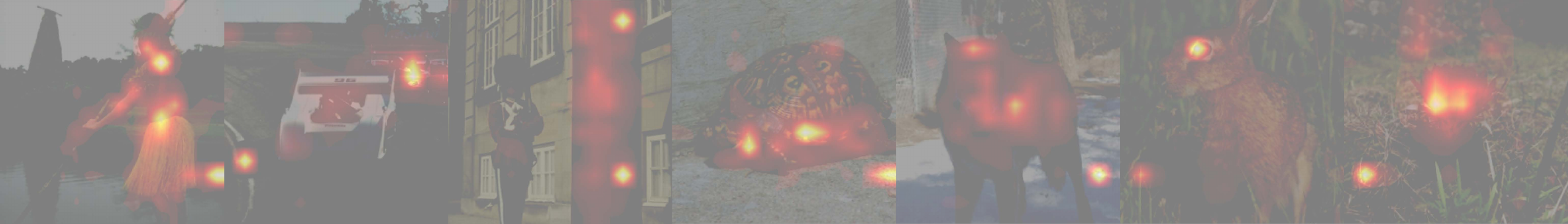} \\
         no local crops, ${\rm mIoU} = 0.34$ \\
         \includegraphics[width=\curWidth]{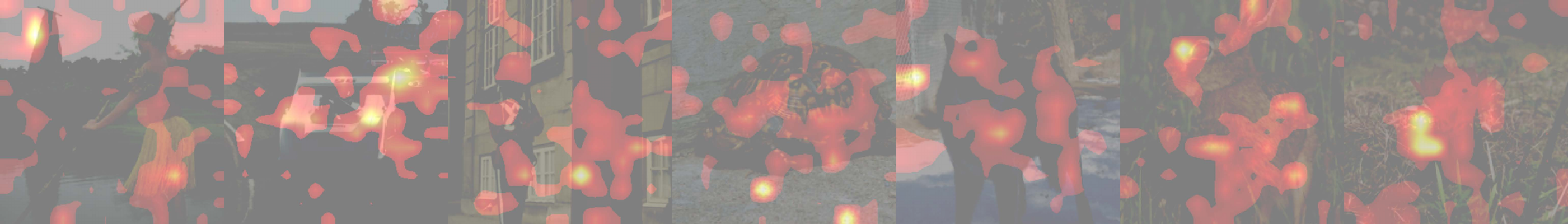} \\
    \end{tabular}
    \caption{Effect of attentive tokenization on the visual grounding of tokens. Training with local crops yields much better token--object alignment than training with global crops only. Attention maps from the 10th layer of \methodName-B; both models embed only the global crop.}
    \label{fig:attn_maps_no_local_crops}
\end{figure}

\textbf{Attentive tokenization.} Next, we evaluate the influence of the presence of local crops during the training. To this end, we train \methodName-B with and without local crops. The variant trained without local crops lacks recurrent message refinement and effectively reduces to a standard encoder-decoder architecture with a discrete bottleneck.

First, we evaluate both variants on a visual recognition benchmark using the same cropping policy at inference time, which encodes only the global crop. Although local crops are not used to construct the latent message at test time, the model trained with the attentive communication bias produces better structured latent messages as reflected with the probing accuracy shown in Table \ref{tab:no_local_ablation}. 

Next, we analyze token attention maps following \citet{duggal2024adaptive}. Using CSSD~\citep{yan2013hierarchical}, we binarize each attention map by retaining the top $Q\%$ of values and select, per image, the token whose map attains the highest IoU~\citep{everingham2010pascal} with the ground-truth mask. All variants are frozen ImageNet-pretrained models evaluated on all 200 CSSD images with the same single-global-crop policy (10th-layer attention at $t=0.1$, $Q=30$); as the mask selects the best token, this is an oracle diagnostic rather than a segmentation method (see Appendix~\ref{app:cssd_protocol} for details). Figure~\ref{fig:attn_maps_no_local_crops} compares these token attention maps for \methodName-B and a variant trained without local crops. \methodName produces attention maps tightly aligned with objects, achieving a mean mIoU of 0.53, whereas the ablated model exhibits diffuse attention across the image and reaches only 0.34 mIoU (95\% image-bootstrap interval of the difference: $[0.175, 0.204]$). This alignment emerges from our attentive tokenization pipeline, indicating that while distillation enforces semantics, tokenization pipeline is key to structuring messages. Table~\ref{tab:srepa_ablation} reports the same diagnostic for the SREPA ablation: removing SREPA costs 10.65 IN100 points but only 0.020 mIoU, whereas removing local crops costs 1.97 points but 0.189 mIoU. Tables~\ref{tab:srepa_ablation} and~\ref{tab:no_local_ablation} thus show complementary signatures: SREPA mostly governs the category content of the message, whereas recurrent local crops mostly govern where it is stored. Additional results are provided in Appendix~\ref{app:attn_maps}.

\textbf{Cropping policy.} Table~\ref{tab:policy_ablation} compares different cropping policies. A single global crop yields competitive average performance across tasks while incurring the lowest test-time cost (the number of crops). Based on this observation, we adopt global-only cropping as the default for the remainder of our probing experiments unless stated otherwise. Importantly, \methodName also naturally supports test-time scaling: adding local crops provides modest gains on compositional generalization and inter-object relations, suggesting that local processing can benefit certain benchmarks. Notably, all differences across cropping policies arise purely at test-time, without retraining the model.

\begin{table*}[t]
    \centering
    \footnotesize
    \caption{Effect of the cropping policy with \methodName-B. A single global crop gives the best trade-off between performance and test-time cost, while some tasks benefit from additional local crops.
    }
    \label{tab:policy_ablation}
    \begin{tabular}{ccc|ccc}
    \toprule
        & & & \multicolumn{1}{c}{IN100} & MSCOCO & VG \\
    Global crop & \# Crops & Ordering &
    top-1 $\uparrow$ &
    top-5 $\uparrow$ &
    top-1 $\uparrow$ \\
    \midrule
    \checkmark & 1  & --               
         & \best{82.91} & \second{41.46} & 52.11 \\

    \midrule
    \checkmark & 10 & Random           
         & \second{82.30} & 40.21 & 52.12 \\
    -          & 9  & Random           
         & 74.45 & 38.92 & 54.96 \\

    \midrule
    \checkmark & 10 & Raster-scan
         & 82.22 & \best{41.83} & \best{55.78} \\
    -          & 9  & Raster-scan
         & 80.62 & 38.61 & 51.10 \\


    \midrule
    \checkmark & 3  & Adaptive         
         & 82.09 & 40.68 & \second{55.00} \\
    --          & 3  & Adaptive         
         & 74.96 & 32.70 & 50.22 \\
    \bottomrule
    \end{tabular}
\end{table*}





\begin{figure*}[t]
    \centering
    \includegraphics[width=\cropsfigwidth\linewidth, clip, trim=0cm 9.5cm 0cm 0cm]{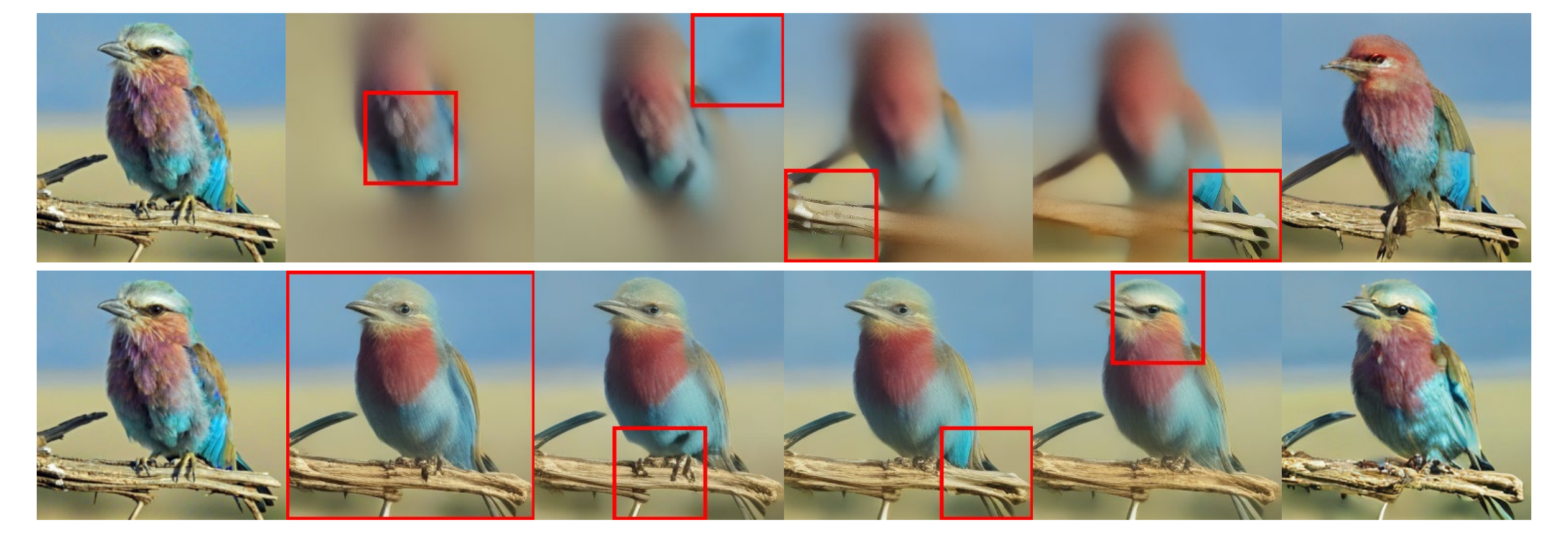}
    \caption{The difference between the \textit{adaptive} (top) and \textit{global+adaptive} (bottom) cropping policies. In both cases \methodName aggregates the crops of the input image (the leftmost column) into the latent message and decodes it to obtain the reconstructed image (the rightmost column, 10 NFE with ${\rm CFG}=7.5$). The columns in-between show which crops are selected together with immediate single step reconstructions (1 NFE with ${\rm CFG}=1.0$).} 
    \label{fig:global_vs_local}
\end{figure*}

\subsection{Quantitative Results}\label{sec:quantitative}

We evaluate \methodName against relevant prior work in the 1D tokenization domain (see Table~\ref{tab:main_comparisons}). For a fair comparison, we consider only unsupervised methods that do not rely on additional paired data from other modalities, such as text. We report results on our semantic probing test suite along with rFID~\cite{heusel2017gans} and PSNR~\cite{huynh2008scope} on ImageNet1k~\cite{deng2009imagenet} validation set. All methods, if not stated otherwise, were pretrained on ImageNet1k~\cite{deng2009imagenet}. \methodName consistently outperforms prior work on semantic tasks, including FlexTok, which also distills DINOv2 features through REPA.

The semantic probes measure what the complete message encodes, but not whether individual tokens are tied to objects. To compare tokenizers on this axis, Appendix~\ref{app:coco_localization} scores the attention map of every individual token against the object masks of multi-object COCO images, with the extraction layer calibrated separately for each tokenizer. \methodName-L localizes 36.52\% of the large objects, 14.78 points more than FlexTok-d18-d18, and spreads its best matches over twice as many tokens. Importantly, token-level localization does not rely on SREPA: \methodName-B trained without SREPA, whose semantic supervision then matches that of FlexTok (DINOv2 REPA only), localizes objects as often as FlexTok-d18-d18 (23.48\% versus 21.74\%) with a fifth of its parameters, whereas removing the recurrent local crops lowers localization to 17.39\%.

Similarly to prior work~\cite{bachmann2025flextok}, we observe that scaling the model from B to L improves both reconstruction and representation, and further scaling the model from L to XL allocates the additional capacity to aid reconstruction, while diminishing the semantics. 

Among the tokenizers evaluated under our frozen-token protocol, \methodName attains the best semantic scores while trading off some reconstruction quality. This representation-reconstruction trade-off is intentional: our work argues that improving the semantics  of token sequences is important for better performance on understanding tasks. The trade-off is not fixed: scaling from B to L improves rFID from 11.06 to 3.67 while all three semantic scores improve, and enlarging the bottleneck improves PSNR at unchanged probing accuracy (Figure~\ref{fig:voc_size} in the Appendix). At the same time, we believe the reconstruction fidelity of \methodName can be further improved with additional training. Many prior methods use multi-stage training pipelines, including decoder fine-tuning to improve reconstruction quality~\citep{yu2024image, chang2023muse}. Importantly, weaker reconstruction does not imply weaker spatial understanding. To demonstrate this, we additionally ran  segmentation probing over a subset of MSCOCO, where SelfTok-512 achieves 0.44 IoU, while \methodName-L achieves 0.53 IoU (identical frozen-token cross-attention decoder predicting the binary foreground mask; see Appendix~\ref{app:segmentation_probe} for details). This shows that lower rFID, e.g., 0.70 for SelfTok-512, does not necessarily translate to better dense prediction performance.


\begin{table*}[t]
    \footnotesize
    \centering
    \caption{We evaluate \methodName and several baseline 1D tokenizers on our test suite comprising three representative benchmarks described in Section~\ref{sec:experiments}. \methodName outperforms prior work on semantic probing. $^\dagger$: trained on larger data mixtures.}
    \label{tab:main_comparisons}
    \setlength{\tabcolsep}{2.5pt}
    \begin{tabular}{@{}lcccrrrrr@{}}
    \toprule
       & & & & \multicolumn{2}{c}{IN1k} & IN100 & MSCOCO & VG \\
       Method & \#params & msg len. & voc. & rFID$\downarrow$ & PSNR$\uparrow$ & top-1$\uparrow$ & top-5$\uparrow$ & top-1$\uparrow$ \\
       \midrule
       TiTok-L~\citep{yu2024image} & 614M & 32 & $2^{12}$ & 2.21 & 15.60 & 17.26 & 6.22 & 26.06 \\
       TiTok-B~\citep{yu2024image} & 172M & 64 & $2^{12}$ & 1.70 & 16.80 & 19.43 & 12.64 & 27.31 \\
       TiTok-S~\citep{yu2024image} & 44M & 128 & $2^{12}$ & 1.71 & 17.52 & 19.47 & 6.30 & 26.81 \\
       ALIT~\citep{duggal2024adaptive} & 229M & 256 & $2^{10}$ & 7.39 & 19.76 & 29.43 & 8.56 & 32.14 \\
       
       FlexTok{\tiny (d12-d12)}~\citep{bachmann2025flextok} & 254M & 256 & 64k & 4.20 & 18.41 &  80.25 & 38.17 & 53.75 \\
       FlexTok{\tiny (d18-d18)}~\citep{bachmann2025flextok} & 860M & 256 & 64k & 1.61 & 18.59 &  81.54 & 38.58 & 54.46 \\
       FlexTok{\tiny (d18-d28)}~\citep{bachmann2025flextok} & 2.5B & 256 & 64k & 1.45 & 18.46 & 80.93 & 39.14 & 54.35 \\
       SelfTok~\cite{wang2025selftok}$^\dagger$ & 2.17B & 512 & $2^{15}$ & \second{0.70} & \second{24.14} & 40.64 & 17.91 & 37.15 \\
       SelfTok~\cite{wang2025selftok}$^\dagger$ & 2.17B & 1024 & $2^{16}$ & \best{0.54} & \best{26.30} & 35.90 & 17.32 & 36.69 \\
       \midrule
       \methodName-B (ours) & 174M & 256 & 64k & 11.06 & 17.75 &  82.91 & \second{41.46} & 52.11 \\
       \methodName-L (ours) & 610M & 256 & 64k & 3.67 & 17.81 & \best{85.80} & \best{45.31} & \best{56.42} \\
       \methodName-XL (ours) & 900M & 
       256 & 64k & 3.50 & 17.83 & \second{84.69} & 39.45 & \second{55.61} \\
       \bottomrule
    \end{tabular}
\end{table*}

\subsection{Qualitative Results}

\begin{figure*}[t]
    \centering
    \newcommand{\curWidth}{\compfigwidth\linewidth}
    \begin{tabular}{@{}c@{}}
        \includegraphics[width=\curWidth]{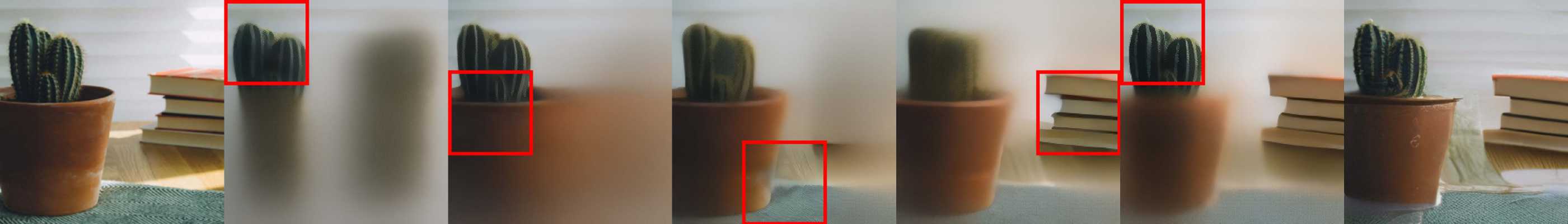} \\
        \includegraphics[width=\curWidth]{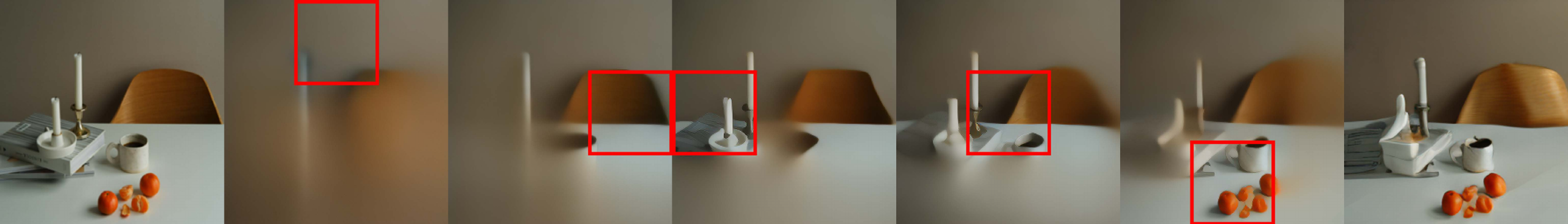} \\
        \includegraphics[width=\curWidth]{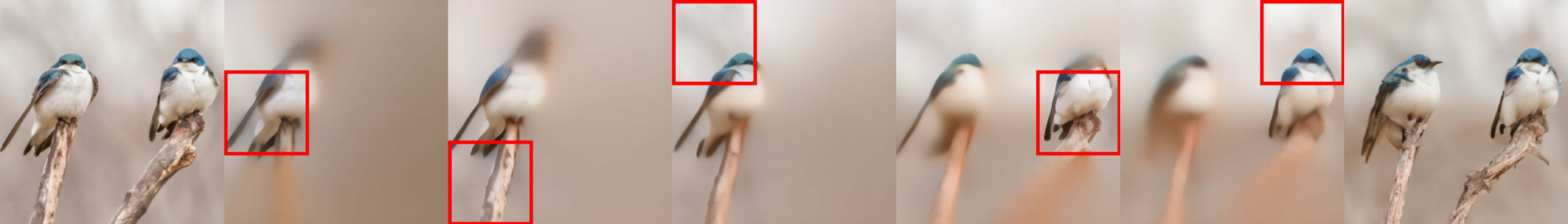} \\
    \end{tabular}
    \caption{The way \methodName adds information to the latent message is inherently compositional. }
    \label{fig:composition}
\end{figure*}

\noindent\textbf{Evolution of Uncertainty.} One way to visually probe the information encoded in the latent messages is via reconstructing the images. However, in the case when only local crops are used to encode the message, 1--2 crops may not be enough to reliably reconstruct the image. Since \methodName is a generative model of images given the message, it will hallucinate the content in the unobserved regions. However, the flow matching framework guarantees that at $t = 0$ the optimal velocity field is the one that points to the mean of the data that shares the same conditioning. Therefore, decoding messages with a single denoising step results in images that have sharp details in the recently observed regions and blurry content in the regions where the model is not certain about the scene. Figures~\ref{fig:global_vs_local} (top) and \ref{fig:composition} demonstrate how the information in the latent message is gradually refined with more crops aggregated.
This process is also inherently compositional. Figure~\ref{fig:composition} shows how objects are added to the latent messages as soon as observed. More visual results can be found in Appendix~\ref{app:visual_results}.


%% file: sections_arxiv/main_paper/conclusion.tex
\section{Conclusion and Discussion}\label{sec:conclusion}

\methodName opens a path toward more structured and semantically meaningful 1D visual tokenization. By treating tokenization as an iterative communication process, it encourages visual information to be organized across discrete token sequences in a way that better supports semantic grounding, compositionality, and relational reasoning. Future directions are discussed in Appendix~\ref{app:future_work}.

%% file: sections_arxiv/main_paper/acknowledgements.tex
\section*{Acknowledgements}

This work was supported as part of the
Swiss AI Initiative by a grant from the Swiss National Supercomputing Centre (CSCS) under project IDs a144 and a137 on Alps.
Aram Davtyan, Sebastian Stapf and Pablo Acuaviva have been supported by
Swiss National Science Foundation (SNSF) Project 10001278. Yasaman Haghighi has been supported by
Swiss National Science Foundation (SNSF) Project 10003100.

%% file: sections_arxiv/appendix/additional_implementation_details.tex
\section{Additional Implementation Details}\label{app:implementation_details}

In this section, we provide additional implementation details critical for reproducing the results in the paper. The code and the pretrained models will be made publicly available upon publication.

\noindent\textbf{Model architecture.} As noted in the main paper, the architecture of \methodName is based on DiT~\citep{peebles2023scalable}. The only substantial difference is that we use different set of AdaLN~\citep{perez2018film} layers to modulate the image and the message tokens. This is similar to \citep{yang2024cogvideox}, where different modulations were applied to text and video tokens. Other architectural details, such as the number of layers and transformer dimensions can be found in Table~\ref{tab:architecture}.

\noindent\textbf{Unified encoder and decoder.} Our design choice was to avoid encoder-decoder separation and allow the model to autonomously decide which part of the network's capacity to dedicate to which task. This choice mimics the symmetry of the communication game, eases development by removing a degree of freedom in designing the architecture and potentially removes redundancies in the model's parameters. We leave further exploration of such fused architectures, including the analysis of how different tasks are balanced in a single network, to future work.

\noindent\textbf{Training hyperparameters.} All models are trained on full ImageNet1k~\citep{deng2009imagenet} using Adam~\citep{kingma2020method} optimizer with a base learning rate of 3e-4 and square root decay schedule with 5000 iterations of warmup. The B and L variants were trained with batch size of 512 images, while for the XL variant we used batch size of 256 with 2 gradient accumulation steps, resulting in the same effective batch size of 512 images. Our models were trained for 200 epochs on 32 $\times$ GH200 GPUs. We use EMA with decay 0.999. For convenience, all hyperparameters are summarized in Table~\ref{tab:architecture}.

\begin{table}[h]
    \centering
    \begin{tabular}{l|ccc}
    \toprule
         & \methodName-B & \methodName-L & \methodName-XL \\
    \midrule
        \multicolumn{4}{c}{\textit{DiT parameters}}\\
    Depth & 12 & 24 & 28 \\
    Hidden size & 768 & 1024 & 1152 \\
    Heads & 12 & 16 & 16 \\
    MLP ratio & 4 & 4 & 4 \\
    \midrule
        \multicolumn{4}{c}{\textit{REPA/SREPA parameters}} \\
    Representation layer & 4 & 4 & 4 \\
    Number of MLP layers & 3 & 3 & 3 \\
    MLP hidden size & 768 & 768 & 768 \\
    \midrule
        \multicolumn{4}{c}{\textit{Bottleneck parameters}} \\
    Number of message tokens & \multicolumn{3}{c}{256} \\
    Token dimensions & \multicolumn{3}{c}{6} \\
    Vocabulary size & \multicolumn{3}{c}{64000} \\
    FSQ Levels & \multicolumn{3}{c}{[8, 8, 8, 5, 5, 5]} \\
    \midrule
        \multicolumn{4}{c}{\textit{Other training parameters}} \\
    Flow matching time distribution & \multicolumn{3}{c}{logitnormal} \\
    Cropping policy & \multicolumn{3}{c}{randomized} \\
    Mode of number of crops $K_{\rm mode}$ & \multicolumn{3}{c}{1} \\
    Maximum number of crops $K_{\rm max}$& \multicolumn{3}{c}{9} \\
    Number of updates to backpropagate through & \multicolumn{3}{c}{1} \\ 
    $p_{\rm CFG}$ & \multicolumn{3}{c}{0.18} \\
    $p_{\rm G}$ & \multicolumn{3}{c}{0.55} \\
    Optimizer & \multicolumn{3}{c}{Adam~\cite{kingma2020method}} \\
    Betas & \multicolumn{3}{c}{[0.9, 0.999]} \\
    $\varepsilon$ & \multicolumn{3}{c}{1e-8} \\
    Weight decay & \multicolumn{3}{c}{0.0} \\
    Learning rate & \multicolumn{3}{c}{0.0003} \\
    Scheduler & \multicolumn{3}{c}{square root} \\
    Warmup steps & \multicolumn{3}{c}{5000} \\
    Gradient clipping & \multicolumn{3}{c}{1.0} \\
    Epochs & \multicolumn{3}{c}{200} \\
    Batch size & 512 & 512 & 256 \\
    Gradient accumulation steps & 1 & 1 & 2 \\
    EMA decay & \multicolumn{3}{c}{0.999} \\
    \bottomrule
    \end{tabular}
    \caption{Architecture and training details of \methodName variants.}
    \label{tab:architecture}
\end{table}

\noindent\textbf{Training stability.} Despite the recurrent encoding and the combined objective, optimization is straightforward: \methodName uses one shared transformer and one end-to-end loss (Equation~\ref{eq:total_loss}) with a frozen DINOv2 teacher, and there is no discriminator, alternating optimization, or separate decoder fine-tuning stage. The stop-gradient through all but the last message update bounds activation memory, and gradient clipping and EMA are used as listed in Table~\ref{tab:architecture}. The same recipe converged for the B, L and XL variants without per-size tuning.

\noindent\textbf{Cropping policy at training.} During training the input images are of resolution $256\times 256$. The local crops of resolution $96\times 96$ are extracted at random locations within the image, such that the local crops do not go beyond the image domain. The number of crops $K$ is randomly selected from $\{1, \dots, K_{\rm max}\}$. We found that instead of uniformly selecting $K$ it is better to slightly skew its distribution towards 1. More precisely, similarly to \citet{geiping2025scaling}, we use the following sampling scheme:
\begin{align}
    K = \min(\max(1, \xi + 1), K_{\rm max}),
    \;{\rm where} \; \xi \sim {\rm Poisson}(e^\tau), \; {\rm and} \;
    \tau = \ln(K_{\rm mode}) + \varepsilon \cdot 0.5 - 0.125
\end{align}

\noindent\textbf{Flow matching timestamp.} In the flow matching framework~\citep{lipmanflow}, the default choice for the timestamp sampling distribution is uniform on the unit interval $U[0, 1]$. However, in prior work it has been discovered that sometimes it is better to skew the timestamp distribution towards the ends of the interval~\citep{esser2024scaling}. We found that it is better to skew the distribution of the timestamps towards 0. This prioritizes high noise regimes and allows for better training of the encoder. We use the logitnormal distribution with $\sigma^2 = 1$ and $\mu = -1$.

\noindent\textbf{Message handling.} For initializing the message tokens we pick one fixed token from the vocabulary (namely with \texttt{id} 32000) and use it to initialize all tokens in $m_0$. Prior to feeding the message tokens to the DiT, we embed them using a 2 layer MLP with a SiLU~\citep{elfwing2018sigmoid} activation in between. As noted in the main paper, at each message update step, we append the current message with a new one that acts as a buffer to write the aggregated information from the message and the crop to. The tokens in the buffer message are copies of learned embedding.

\noindent\textbf{Position encodings.} Each modality in the input uses its own set of position encodings. The message and the offset tokens are augmented with learned position encodings. The image tokens use 2D sine--cosine position encodings. We define some maximum resolution of the image input (depends on the resolution of the images, the downsampling factor of the VAE and the patch size of the DiT). Each crop only uses the top-left corner of the 2D image position encodings.

\noindent\textbf{Classifier-free guidance.} To smoothen the dependence of the metrics on the CFG strength and allow for high values of the latter, similarly to \cite{bachmann2025flextok}, we use the Adaptive Projected Guidance~\citep{sadat2024eliminating} with the rescaling threshold as $r = 2.5$, parallel component as $\eta = 0$, and momentum as $\beta = -0.5$. 


%% file: sections_arxiv/appendix/additional_evaluation_details.tex
\section{Evaluation Details}\label{app:test_suite}

\subsection{Test Suite}
We evaluate the ability of image tokenizers to encode semantic content using a probing protocol. Images are first tokenized using the evaluated tokenizers, and the resulting image-token sequences are kept frozen. For each benchmark, we train a lightweight probe with identical capacity across tokenizers to ensure a fair comparison. Across all tasks, we apply early stopping based on validation performance and report results from the best-performing checkpoint.

\noindent\textbf{ImageNet100.}
We evaluate category-level semantic information using ImageNet100 as a single-label classification task. Given frozen image tokens, the probe is trained to predict one of 100 object categories using a cross-entropy objective. Performance is measured using top-1 accuracy on the validation set.

\noindent\textbf{MSCOCO.}
We assess compositional generalization using a multi-label classification task derived from MSCOCO. We retain only images containing exactly two object categories and construct pair-disjoint training and validation splits such that object category pairs observed at validation time are never seen during training. Each image is annotated with a 2-hot target vector over 90 categories. The probe is trained with a binary cross-entropy loss, and performance is reported using top-5 accuracy, requiring both ground-truth categories to appear among the top-5 predictions.

\noindent\textbf{Visual Genome.}
We evaluate relational reasoning on Visual Genome, which provides supervision in the form of $(\text{subject}, \text{predicate}, \text{object})$ triplets with a large and open-ended vocabulary. Training and validation splits are constructed to be image-disjoint to avoid visual leakage. To ensure a meaningful probing signal, we filter out relations involving objects that occupy less than 5\% of the image area and restrict the vocabulary to the 150 most frequent subject/object categories and predicates. This preprocessing avoids evaluation failures caused by out-of-vocabulary textual concepts at validation time, allowing the probe’s performance to more faithfully reflect the semantic information encoded in the image tokens rather than limitations of the classifier’s label space. For each image, the probe is trained to distinguish the correct triplet from a set of negative triplets sampled from other images using a contrastive objective. At evaluation time, accuracy is reported based on whether the correct triplet is identified among its associated negatives.

\subsection{Probing Network Design}
All probes operate on frozen image-token sequences and use a shallow Transformer encoder to aggregate information across tokens. Image tokens $\mathbf{m}\in\mathbb{R}^{L\times d}$ are first projected to a shared model dimension, and a learnable \texttt{[CLS]} token is prepended to the sequence. Learned positional embeddings and discrete type embeddings are added to all inputs before applying self-attention. Predictions are obtained from the final \texttt{[CLS]} representation.

For ImageNet100 and MSCOCO, the input sequence consists only of the \texttt{[CLS]} token followed by the projected image tokens. A two-way type embedding distinguishes image tokens from \texttt{[CLS]}. The Transformer output corresponding to \texttt{[CLS]} is passed to a linear classification head, producing either single-label logits (ImageNet100) or multi-label logits (MSCOCO).

For Visual Genome, the probe additionally conditions on a candidate $(\text{subject}, \text{predicate}, \text{object})$ triplet. Subject, object, and predicate text are encoded using a frozen T5 encoder~\citep{raffel2020exploring} and independently projected to the model dimension. These embeddings are appended to the image-token sequence, and a five-way type embedding is used to distinguish image tokens, textual roles, and \texttt{[CLS]}. The Transformer encoder processes the full sequence jointly, and a linear head maps the \texttt{[CLS]} representation to a single logit representing image–triplet compatibility.
\begin{table}[t]
\centering
\begin{tabular}{lcccc}
\toprule
Benchmark & Model dim & Depth & Heads & Global Batch Size \\
\midrule
ImageNet100   & 768 & 2 & 8 & 512 \\
MSCOCO        & 768 & 2 & 8 & 256 \\
Visual Genome & 64 & 2 & 8 & 512 \\
\bottomrule
\end{tabular}
\caption{Probe hyperparameters used for all benchmarks.}
\label{tab:hyperparameters}
\end{table}
\subsection{Hyperparameters}
All probes are trained using the AdamW~\citep{loshchilov2017fixing} optimizer with a fixed learning rate, which we keep constant across benchmarks to ensure comparability. For Visual Genome, we use a smaller model dimension than for ImageNet100 and MSCOCO due to the increased input length from concatenating textual triplet embeddings and the substantially larger effective batch size required for contrastive training with multiple negatives, which together impose higher memory and compute demands. An overview of the hyperparameters is given in Table~\ref{tab:hyperparameters}.

\begin{figure}[h]
    \centering
    \begin{subfigure}[t]{0.32\linewidth}
        \centering
        \includegraphics[width=\linewidth]{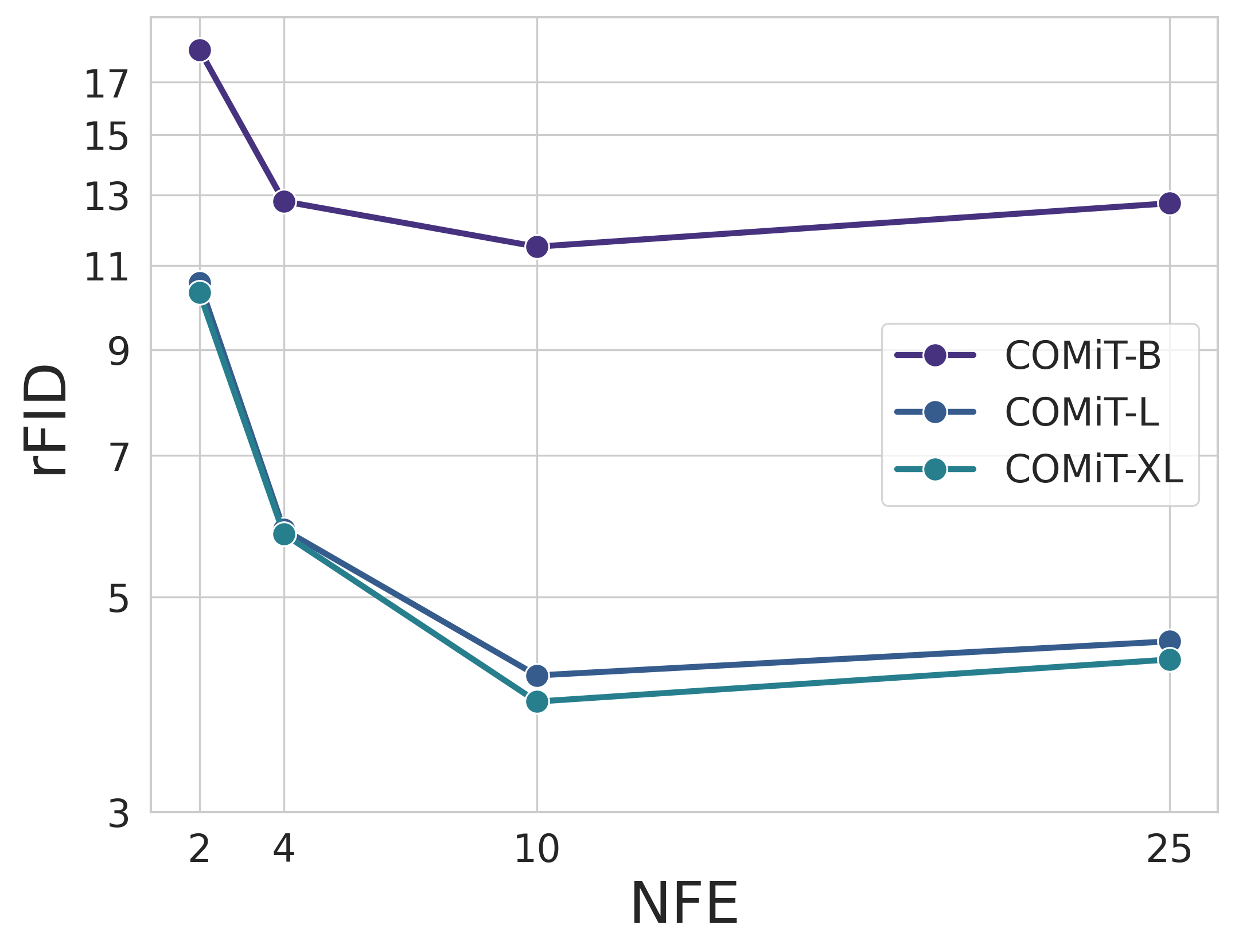}
        \caption{rFID vs. NFE with ${\rm CFG} = 3.0$}
        \label{fig:nfe}
    \end{subfigure}
    \begin{subfigure}[t]{0.32\linewidth}
        \centering
        \includegraphics[width=\linewidth]{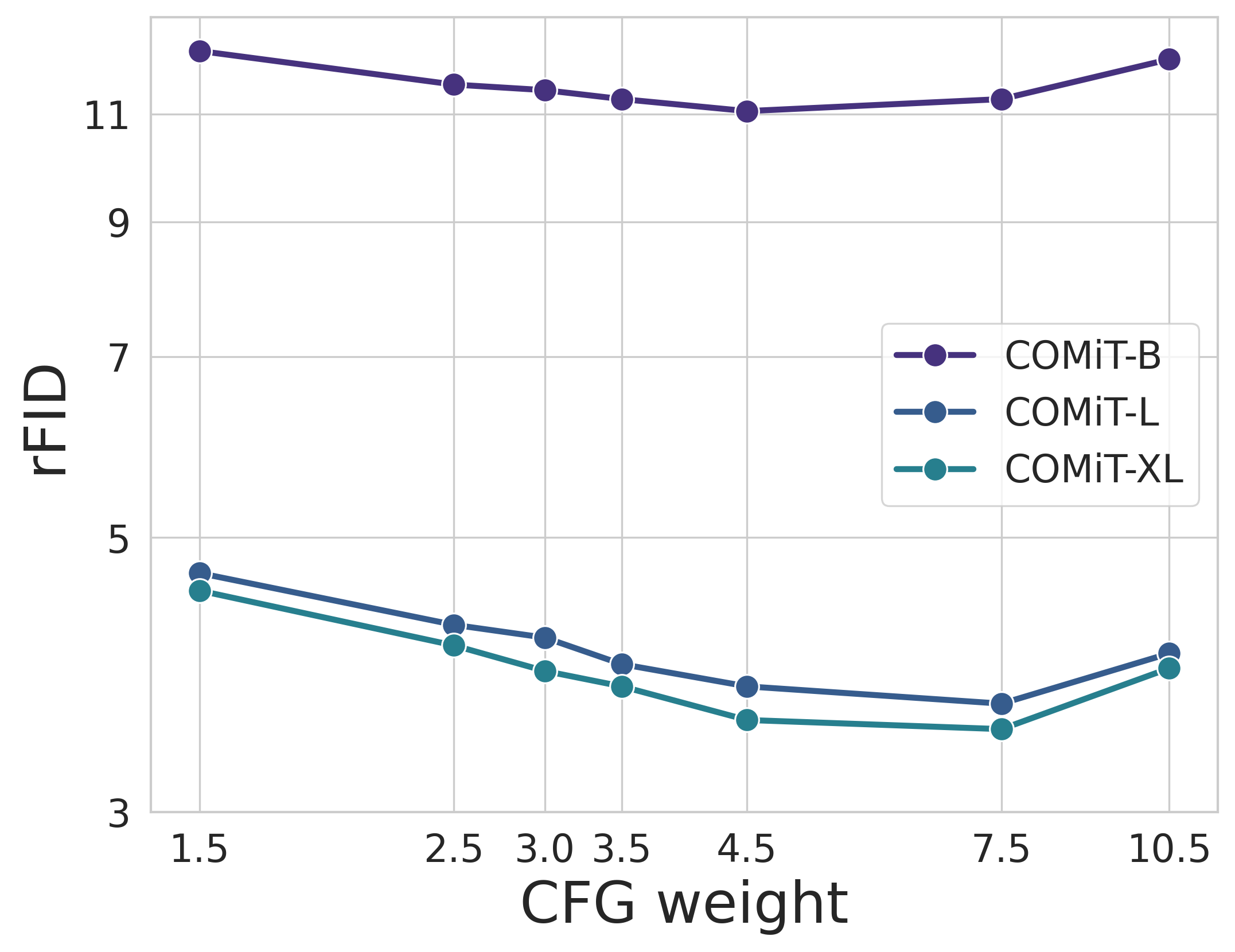}
        \caption{rFID vs. CFG with ${\rm NFE} = 10$}
        \label{fig:cfg}
    \end{subfigure}
    \begin{subfigure}[t]{0.32\linewidth}
        \centering
        \includegraphics[width=\linewidth]{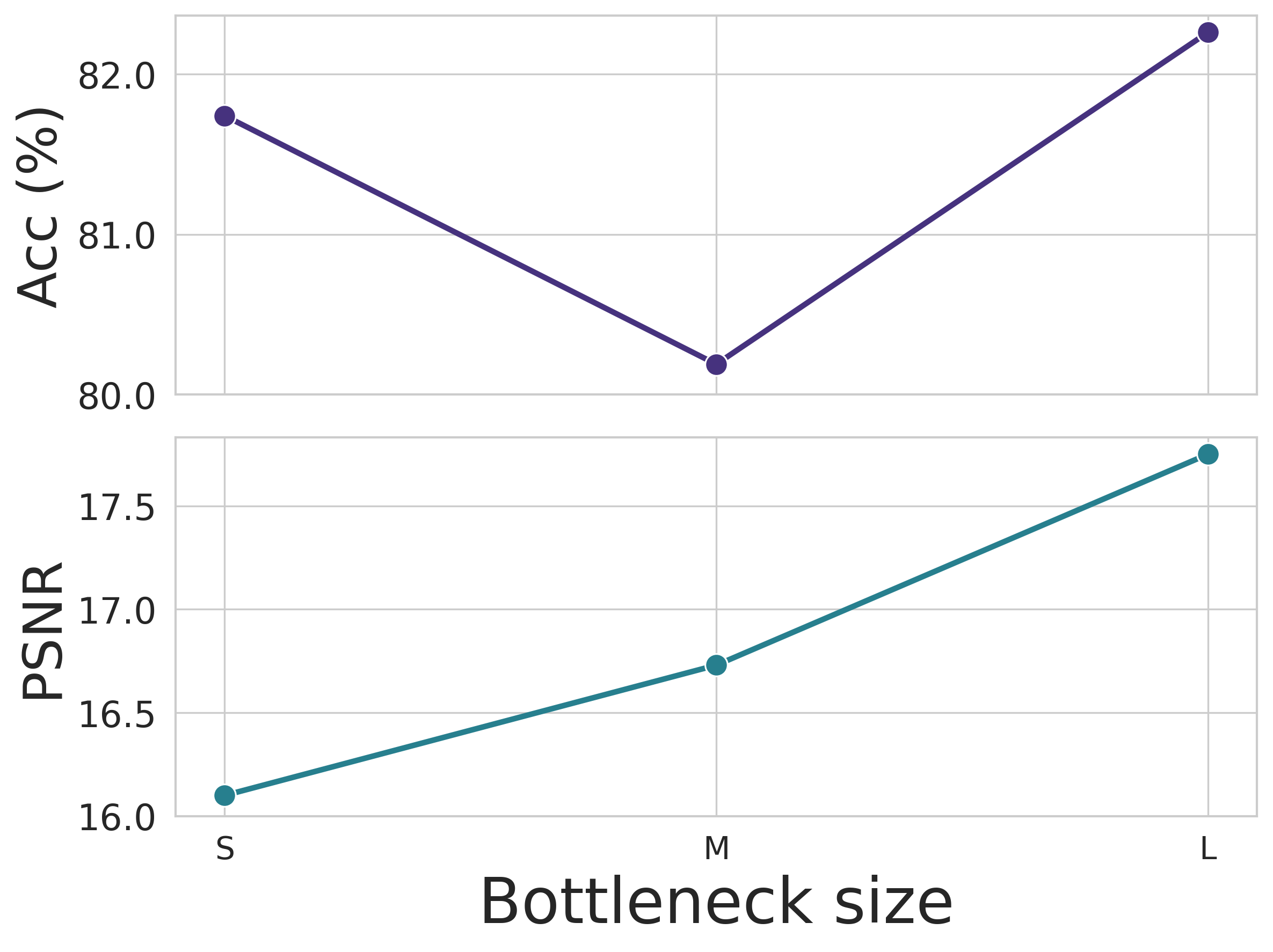}
        \caption{Ablation of the bottleneck size}
        \label{fig:voc_size}
    \end{subfigure}
    \caption{(a) and (b): Ablation of the reconstruction fidelity under different sampling hyperparameters. (c): Ablation of the bottleneck size with \methodName-B.}
    \label{fig:psnr_acc}
\end{figure}

\subsection{Reconstruction.} To select the optimal parameters for the NFE and CFG we start by fixing ${\rm CFG} = 3.0$ and sweep the NFE in $\{2, 4, 10, 25\}$ (see Figure~\ref{fig:nfe}). We find ${\rm NFE} = 10$ to be the optimal number of evaluation steps across all model sizes. We observed that increasing NFE typically leads to sharper details and more salient artifacts, which slightly increases rFID. We then fix ${\rm NFE} = 10$ and sweep CFG in $\{1.5, 2.5, 3.0, 3.5, 4.5, 7.5, 10.5\}$ (see Figure~\ref{fig:cfg}). We observe the rFID curves behave smoothly under the changes in CFG, with ${\rm CFG} = 7.5$ yielding the lowest rFID with \methodName-L and -XL. Therefore, for all our experiments we fix ${\rm NFE} = 10$ and ${\rm CFG} = 7.5$. Notably, scaling \methodName from B to L leads to significant drop in rFID, while further scaling from L to XL only moderately improves the reconstruction quality.

\subsection{Bottleneck size.} We additionally ablate the bottleneck size of \methodName. We train 3 variants of \methodName-B with small (S), medium (M) and large (L) bottleneck sizes, where S corresponds to the message length of 64 tokens from a vocabulary of 1k unique entries, M stay for the message length of 128 tokens and 16k vocabulary size, while L is the default configuration in the main paper with 256 tokens in the messages and 64k tokens in the vocabulary. We then report the classification probing accuracies of these variants along with the reconstruction PSNRs (see Figure~\ref{fig:psnr_acc}). We observe that the accuracy stays within the same range, while PSNR keeps increasing. This may indicate the potential to further improve the reconstruction fidelity of the model without losing the probing capabilities. 

%% file: sections_arxiv/appendix/attn_maps.tex
\section{Analysis of the Attention Maps}\label{app:attn_maps}

In this section, we extend the preliminary analysis of the tokens' attention maps demonstrated in the main paper.

\begin{figure}[t]
    \centering
    \includegraphics[width=0.9\linewidth]{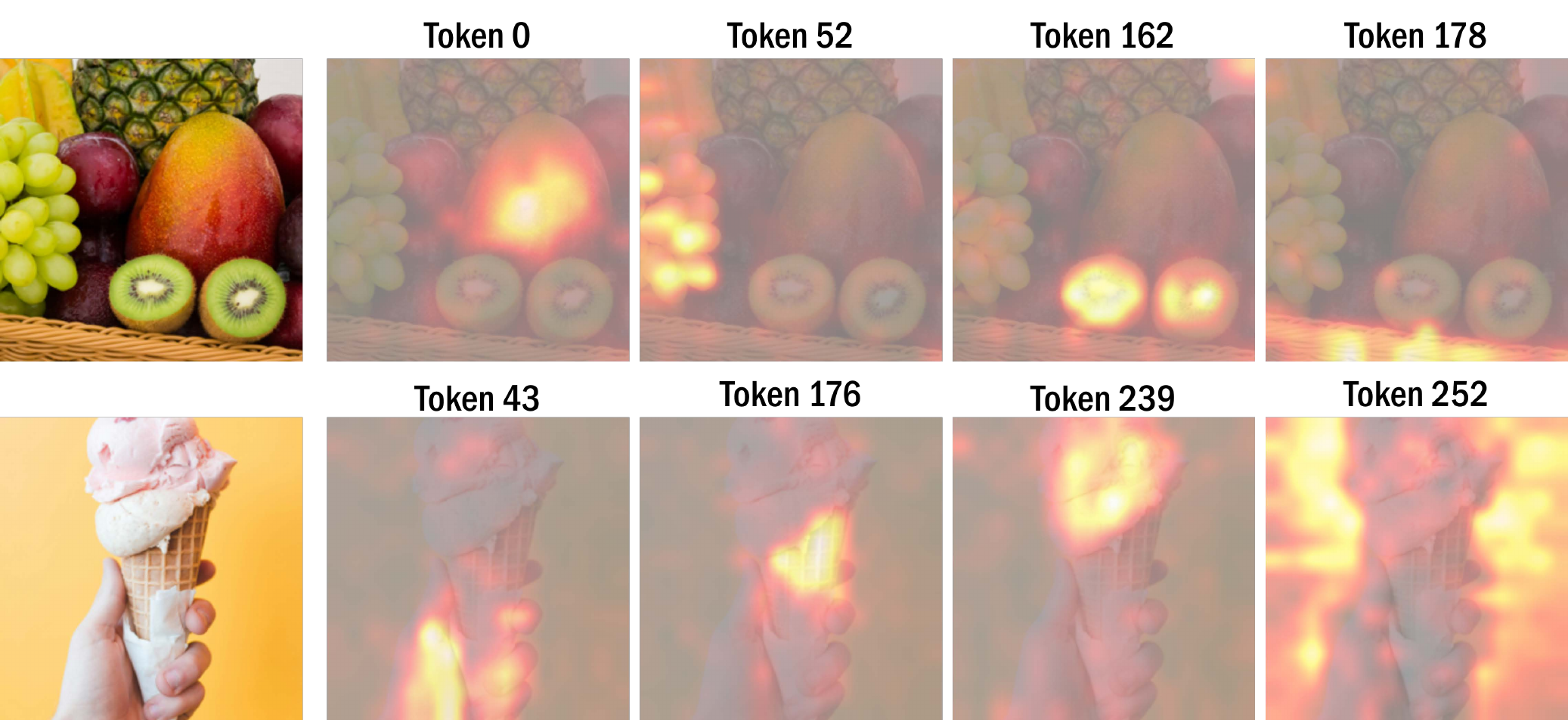}
    \caption{Emergence of object-centric tokens in \methodName. We visualize the attention maps of specific tokens in one of the deep  layers of the network during the decoding stage. One can see that the tokens naturally attend to the objects and their parts.} 
    \label{fig:attn_maps}
\end{figure}

\noindent\textbf{Emergence of Objectness.} Similarly to \citep{duggal2024adaptive}, we study the emergence of object discovery via visualizing the attention maps of message tokens to the image tokens in the 24th layer of \methodName-XL during the decoding process. First, we visually investigate the attention maps and observe that \methodName's tokens tend to correspond to some semantically meaningful regions in images, such as objects or object parts (see Figure~\ref{fig:attn_maps}).

\begin{figure}[t]
    \centering
    \newcommand{\curWidth}{\linewidth}
    \begin{tabular}{@{}c@{}}
        Input images \\
         \includegraphics[width=\curWidth]{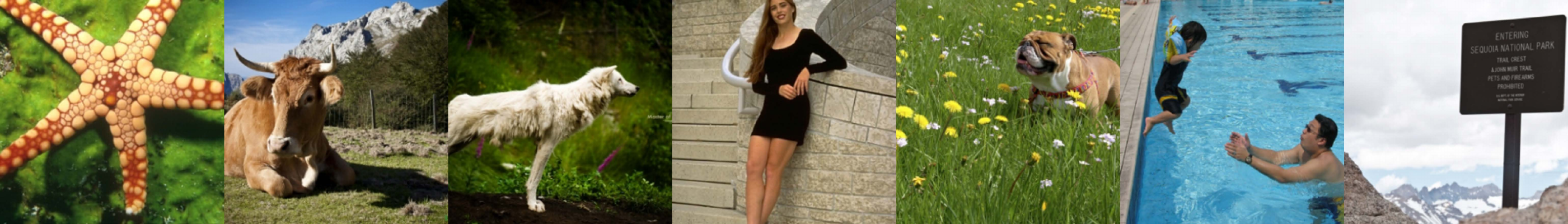} \\
         Ground truth masks \\
         \includegraphics[width=\curWidth]{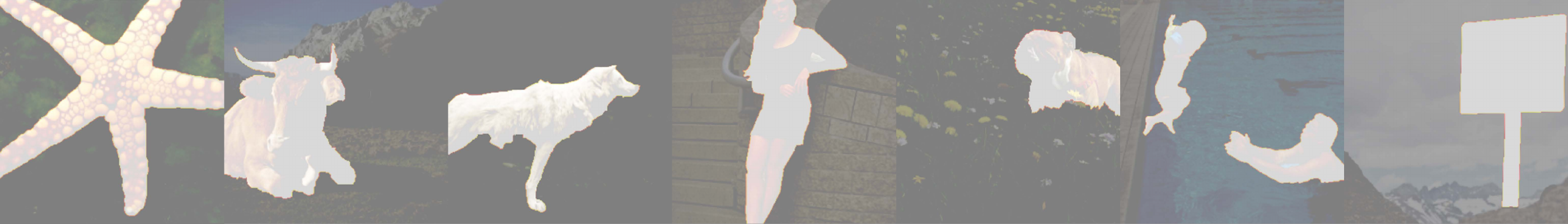} \\
         40\% attention thresholding, ${\rm mIoU} = 0.57$ \\
         \includegraphics[width=\curWidth]{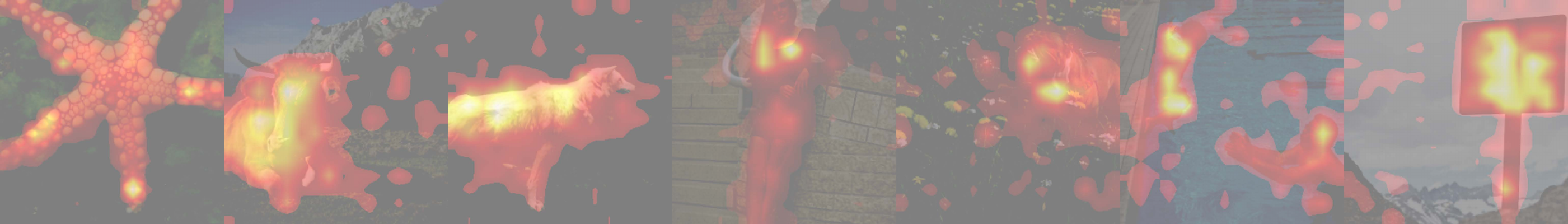} \\
         30\% attention thresholding, ${\rm mIoU} = 0.58$ \\
         \includegraphics[width=\curWidth]{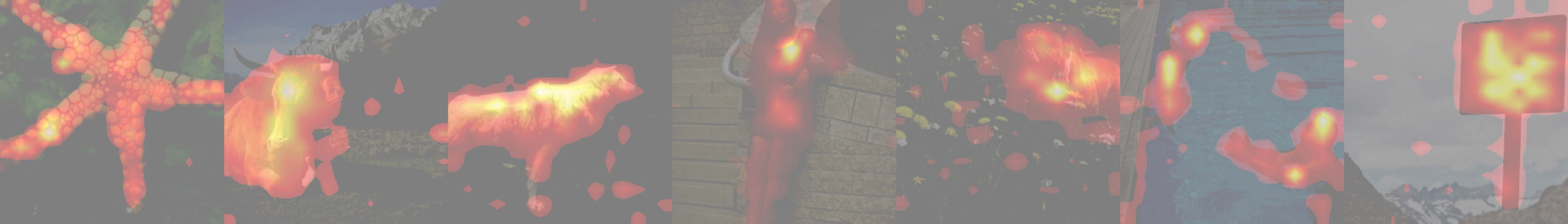} \\
         20\% attention thresholding, ${\rm mIoU} = 0.54$ \\
         \includegraphics[width=\curWidth]{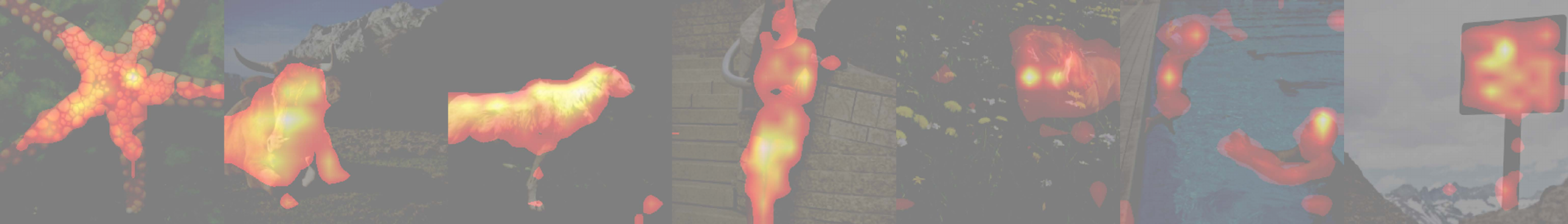} \\
         10\% attention thresholding, ${\rm mIoU} = 0.38$ \\
         \includegraphics[width=\curWidth]{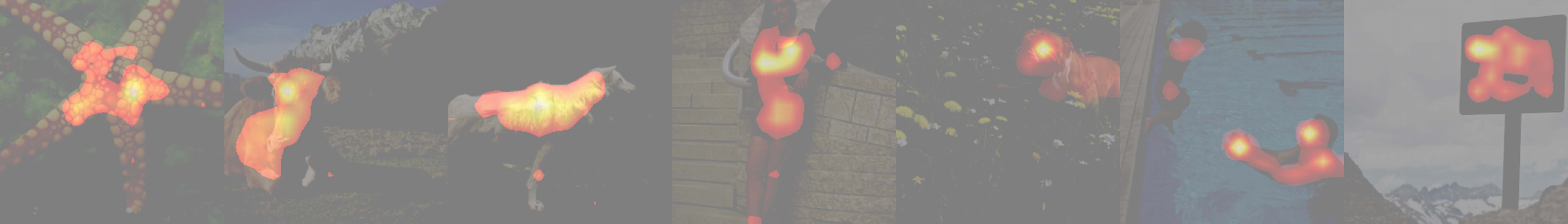} \\
    \end{tabular}
    \caption{Visualization of \methodName-XL's attention maps. The attention maps of the tokens that yield the largest IoUs per sample are shown. With adjusting the thresholding percentage, a remarkable mIoU of 0.58 can be achieved. Note that the model has never seen any segmentation maps or class labels during training.}
    \label{fig:attn_maps_cssd}
\end{figure}

\noindent\textbf{Quantifying Objectness.} To quantify these findings, we follow the same procedure described in the main paper. In Figure~\ref{fig:attn_maps_cssd}, we visualize the attention maps with 10\%, 20\%, 30\% and 40\% thresholding along with the mean IoUs for the whole CSSD dataset~\citep{yan2013hierarchical}. We found that with 30\% thresholding one can obtain 0.58 mIoU, which is considered remarkable for a model that has never seen segmentation maps or object categories during training. The attention maps are extracted from layer 24 at denoising time $t=0.1$ using \methodName-XL.

\begin{figure}
    \centering
    \newcommand{\curWidth}{0.9\linewidth}
    \begin{tabular}{c}
         \includegraphics[width=\curWidth]{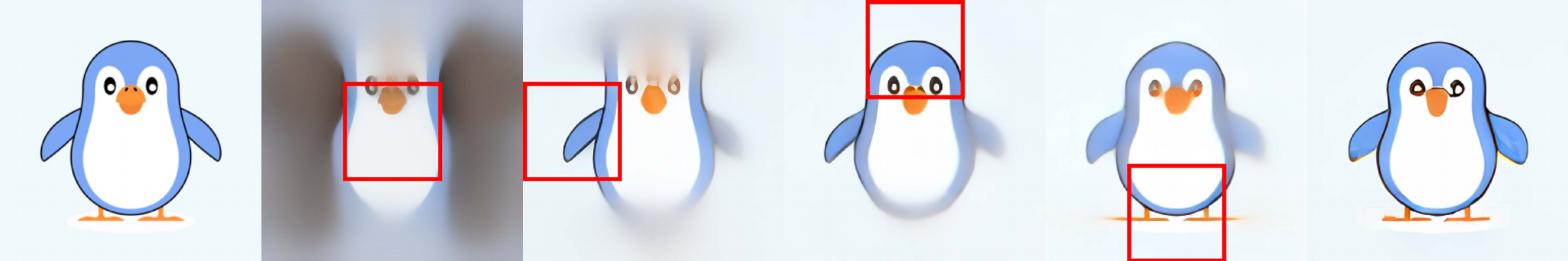} \\
         \includegraphics[width=\curWidth]{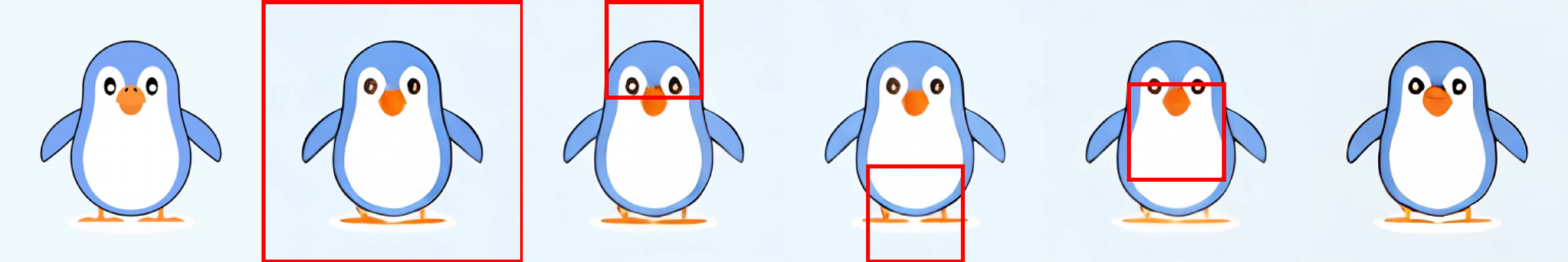} \\
         \includegraphics[width=\curWidth]{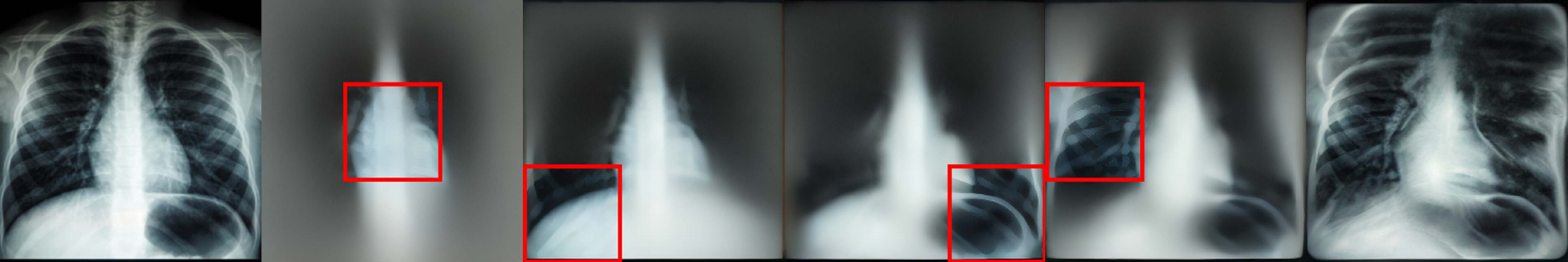} \\
         \includegraphics[width=\curWidth]{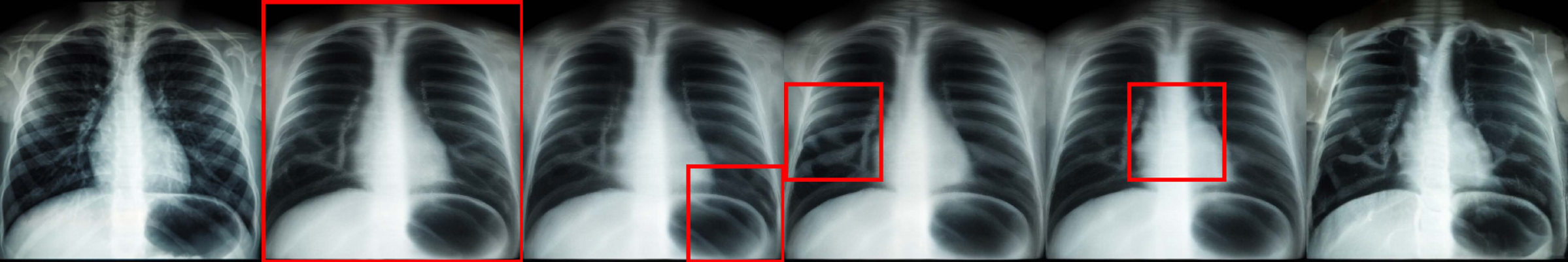} \\
    \end{tabular}
    \caption{Generalization of \methodName to other domains, such as rendered animations or medical images. It can be seen that the model has certain symmetry bias that allows it to reduce the uncertainty about the right hand side of the image when the content on the left hand side is observed. Also notice how the adaptive cropping policy selects the most critical regions for reconstruction.}
    \label{fig:ood}
\end{figure}

%% file: sections_arxiv/appendix/additional_ablations.tex
\section{Additional Ablations and Controls}\label{app:additional_ablations}

This section documents the protocols and additional controls behind the ablations of Section~\ref{sec:ablations} and the experiments in Section~\ref{sec:quantitative}: the exact CSSD localization protocol, a non-attention readout, the robustness to truncated backpropagation, the effect of randomizing the number of crops during training, a cross-tokenizer token-localization study on multi-object COCO images, and the protocol of the segmentation probe.

\subsection{CSSD Token-Localization Protocol}\label{app:cssd_protocol}

All \methodName variants are pretrained on ImageNet1k only and kept frozen. CSSD~\citep{yan2013hierarchical} is used exclusively for evaluation: there is no training, fine-tuning or learned segmentation head, and all 200 images are used. Every variant encodes the image with the same single-global-crop policy. We then decode the message at denoising time $t=0.1$ and extract the message-to-image attention maps of the 10th layer of \methodName-B (the 24th layer of \methodName-XL in Appendix~\ref{app:attn_maps}), averaged over heads. Each token's map is binarized by retaining its top $Q=30\%$ locations, the IoU with the ground-truth mask is computed for every token separately, and the best token is selected per image. The mean of these best IoUs over the dataset is the reported mIoU. Because the ground-truth mask is used to select the token, the score is an oracle localization diagnostic of individual attention maps and not the output of a segmentation system, and it does not combine information across tokens through any learned readout. Table~\ref{tab:cssd_three_way} extends Tables~\ref{tab:srepa_ablation} and~\ref{tab:no_local_ablation} from Section~\ref{sec:ablations} with the corresponding 95\% image-bootstrap intervals, which resample the 200 images and hence quantify dataset uncertainty for these checkpoints.

\begin{table}[h]
    \centering
    \footnotesize
    \caption{Three-way ablation of \methodName-B on IN100 (attention probe) and CSSD token localization. All variants report results for a single global crop at test time. The last column is the paired full-minus-variant difference in mIoU with its 95\% image-bootstrap interval.}
    \label{tab:cssd_three_way}
    \begin{tabular}{@{}lcccc@{}}
    \toprule
    Training variant & IN100 top-1$\uparrow$ & CSSD mIoU$\uparrow$ & 95\% interval & Paired difference \\
    \midrule
    Full & \best{82.91} & \best{0.527} & $[0.509, 0.546]$ & -- \\
    No SREPA & 72.26 & 0.507 & $[0.490, 0.525]$ & $+0.020 \; [+0.012, +0.028]$ \\
    No local crops & 80.94 & 0.338 & $[0.324, 0.352]$ & $+0.189 \; [+0.175, +0.204]$ \\
    \bottomrule
    \end{tabular}
\end{table}

\subsection{Affine Readout Controls}\label{app:affine_probe}

The attention probe of Appendix~\ref{app:test_suite} is shared by all tokenizers in Table~\ref{tab:main_comparisons}, so it does not give \methodName a method-specific advantage. As a complementary check that does not rely on an attention-based head, we also use the simplest position-sensitive readout: the frozen message is flattened in token order and mapped by a single affine layer to the class logits (153,700 parameters for a $256 \times 6$ message on IN100), with no attention and no nonlinear hidden layer. All affine probes use AdamW with a fixed learning rate of $0.01$, no weight decay, batch size 1,024, at most 30 epochs with early stopping (patience 5) on the validation split, and five probe seeds. Each tokenizer is a single pretrained checkpoint, so the reported standard deviations cover probe initialization only. The readout is deliberately position-sensitive: mean pooling the six-dimensional token coordinates before the affine map removes nearly all accessible category information (6.42\% IN100 top-1 for \methodName-B in a single run) and is therefore not an informative comparison.

We apply this protocol to \methodName and to prior tokenizers whose flattened message has the same dimensionality of 1,536 (Table~\ref{tab:affine_cross}). Such a position-specific readout has a different bias than the attention probe: it favors tokenizers that concentrate semantic information at fixed positions, as FlexTok does by construction. Indeed, all FlexTok variants exceed \methodName-B under this readout, whereas \methodName-L exceeds every evaluated baseline. The cross-model result at L scale therefore does not depend on the attention probe.

\begin{table}[h]
    \centering
    \footnotesize
    \caption{Single affine readout over the flattened frozen message of tokenizers with a flattened message dimensionality of 1,536 (IN100 top-1, mean $\pm$ sample SD over five probe seeds; \methodName encodes a single global crop).}
    \label{tab:affine_cross}
    \begin{tabular}{@{}lcc@{}}
    \toprule
    Tokenizer & Message shape & IN100 top-1$\uparrow$ \\
    \midrule
    TiTok-S (128 tokens)~\citep{yu2024image} & $128 \times 12$ & $16.60 \pm 0.14$ \\
    FlexTok{\tiny (d12-d12)}~\citep{bachmann2025flextok} & $256 \times 6$ & $72.07 \pm 0.22$ \\
    FlexTok{\tiny (d18-d18)}~\citep{bachmann2025flextok} & $256 \times 6$ & $74.32 \pm 0.14$ \\
    FlexTok{\tiny (d18-d28)}~\citep{bachmann2025flextok} & $256 \times 6$ & $72.76 \pm 0.38$ \\
    \methodName-B (ours) & $256 \times 6$ & $71.66 \pm 0.25$ \\
    \methodName-L (ours) & $256 \times 6$ & $\mathbf{76.45 \pm 0.28}$ \\
    \bottomrule
    \end{tabular}
\end{table}

\subsection{Robustness to Truncated Backpropagation}\label{app:heuristic_robustness}

Section~\ref{sec:greedy_use_of_tokens} stops gradients through all but the last message update. Table~\ref{tab:heuristic_robustness} evaluates an additional \methodName-B checkpoint under the CSSD protocol of Appendix~\ref{app:cssd_protocol} that propagates gradients through up to the last four local-crop updates (depending on how many crops were sampled). Token--object localization persists, so the effect reported in Section~\ref{sec:ablations} does not depend on the stop-gradient depth; the default is primarily a memory-saving choice. Since the two checkpoints are separate training runs with different seeds, the comparison serves as an indicator of robustness.

\begin{table}[h]
    \centering
    \footnotesize
    \caption{CSSD token-localization mIoU of \methodName-B trained with the default stop-gradient heuristic and with gradients through up to four message updates (single global crop at test time; 95\% image-bootstrap intervals).}
    \label{tab:heuristic_robustness}
    \begin{tabular}{@{}lcc@{}}
    \toprule
    Training setting & CSSD mIoU$\uparrow$ & 95\% interval \\
    \midrule
    One backpropagated update (default) & 0.527 & $[0.509, 0.546]$ \\
    Up to four backpropagated updates & 0.566 & $[0.545, 0.587]$ \\
    \bottomrule
    \end{tabular}
\end{table}

\subsection{Randomized Number of Crops}\label{app:crop_randomization}

Section~\ref{sec:greedy_use_of_tokens} argues that with a fixed number of crops during training the model would pre-allocate parts of the message to the crops it expects to follow, whereas a randomized number of crops encourages it to use the available tokens greedily. We test this with three \methodName-B checkpoints: the default model (randomized number of crops, one backpropagated update), the checkpoint of Appendix~\ref{app:heuristic_robustness} (randomized number of crops, up to four backpropagated updates), and a checkpoint trained with a fixed number of four local crops (one backpropagated update).

\noindent\textbf{Token usage.} Each checkpoint tokenizes the ImageNet100 validation set and a class-balanced subset of the training set (200 images per class), once with the global crop alone and once with the nine local crops of Section~\ref{sec:cropping_policies} in raster-scan order, and the message is recorded after every crop. To measure how much image-specific information a message position $i$ carries, we compute
\begin{align}
    H_i = -\sum_{j=1}^{6} \sum_{l=1}^{L_j} p_{i,j}(l) \log_2 p_{i,j}(l),
\end{align}
where $L_j \in \{8, 5\}$ is the number of FSQ levels of coordinate $j$ (Table~\ref{tab:architecture}) and $p_{i,j}(l)$ is the fraction of the 5,000 validation images in which coordinate $j$ of token $i$ takes level $l$. $H_i$ is 0 if the position holds the same code in every image and at most $\sum_j \log_2 L_j = 15.97$ bits. Since the joint entropy of a code cannot exceed the sum of its coordinate entropies, a position with $H_i < 8$ bits behaves as if it used fewer than $2^8 = 256$ equally frequent codes out of the 64,000 codes of the vocabulary; we call such positions under-used. Table~\ref{tab:crop_randomization_usage} shows the pre-allocation described in Section~\ref{sec:greedy_use_of_tokens}. After its first local crop, the model trained with a fixed number of crops leaves 36.7\% of the message positions under-used (37.1\% with a random crop order) and fills them during the next two crops (3.5\% after the third), whereas the randomized models leave at most 5.9\% of the positions under-used after any number of crops. After the global crop, none of the models leaves more than 1\% of the positions under-used. These differences do not depend on the 8-bit threshold: after the first local crop, the fixed-$K$ model has 19.1\% and 50.0\% of its positions below 4 and 12 bits, respectively, compared with at most 3.5\% and 12.1\% for the randomized models.

\begin{table}[h]
    \centering
    \footnotesize
    \caption{Fraction of under-used message positions (\%, $H_i < 8$ bits) after the global crop or after the first local crops (raster-scan order unless stated otherwise).}
    \label{tab:crop_randomization_usage}
    \begin{tabular}{@{}lccc@{}}
    \toprule
    & \multicolumn{2}{c}{Randomized number of crops} & Fixed number of crops \\
    Message after & one update & up to four updates & one update \\
    \midrule
    Global crop & 0.0 & 0.4 & 0.8 \\
    1 local crop & 4.7 & 5.5 & 36.7 \\
    2 local crops & 4.7 & 3.5 & 9.4 \\
    3 local crops & 5.1 & 3.5 & 3.5 \\
    4 local crops & 3.5 & 4.7 & 3.5 \\
    1 local crop, random order & 4.7 & 5.9 & 37.1 \\
    \bottomrule
    \end{tabular}
\end{table}

\noindent\textbf{Semantic probing.} The affine readout of Appendix~\ref{app:affine_probe} (five seeds) and a two-layer attention probe with the architecture of Appendix~\ref{app:test_suite} (three seeds) are trained on the training subset and evaluated on the full validation set. Because of the smaller training set, absolute accuracies are lower than, and not directly comparable to, those of Table~\ref{tab:srepa_ablation}. Table~\ref{tab:crop_randomization_probe} shows that the randomized models also make class information more accessible. Both exceed the fixed-$K$ model with the global crop and after every number of local crops. The gap persists at four crops, the training length of the fixed-$K$ model, and after the global crop, where the fixed-$K$ model leaves fewer than 1\% of the positions under-used, so the benefit of randomization for probing goes beyond the greedy use of the first crops. The randomized number of crops is the only setting in which both randomized checkpoints differ from the fixed-$K$ one, which makes the randomization critical for both token usage and probing.

\begin{table}[h]
    \centering
    \footnotesize
    \setlength{\tabcolsep}{3pt}
    \caption{ImageNet100 top-1 accuracy (\%) of probes trained on 200 images per class, mean $\pm$ std over probe seeds. R1 and R4: randomized number of crops with one and up to four backpropagated updates; F1: fixed number of four crops with one backpropagated update.}
    \label{tab:crop_randomization_probe}
    \begin{tabular}{@{}lcccccc@{}}
    \toprule
    & \multicolumn{3}{c}{Affine readout (5 seeds)} & \multicolumn{3}{c}{Attention probe (3 seeds)} \\
    Message after & R1 & R4 & F1 & R1 & R4 & F1 \\
    \midrule
    Global crop & 56.08\,{\scriptsize$\pm$0.28} & 63.41\,{\scriptsize$\pm$0.35} & 46.78\,{\scriptsize$\pm$0.46} & 74.71\,{\scriptsize$\pm$0.54} & 77.09\,{\scriptsize$\pm$0.50} & 65.40\,{\scriptsize$\pm$1.68} \\
    1 local crop & 26.44\,{\scriptsize$\pm$0.43} & 29.49\,{\scriptsize$\pm$0.37} & 16.90\,{\scriptsize$\pm$0.15} & 27.55\,{\scriptsize$\pm$1.50} & 30.56\,{\scriptsize$\pm$0.34} & 18.53\,{\scriptsize$\pm$0.75} \\
    2 local crops & 41.54\,{\scriptsize$\pm$0.42} & 46.12\,{\scriptsize$\pm$0.32} & 29.64\,{\scriptsize$\pm$0.36} & 46.93\,{\scriptsize$\pm$1.15} & 49.27\,{\scriptsize$\pm$0.48} & 36.54\,{\scriptsize$\pm$1.52} \\
    4 local crops & 52.69\,{\scriptsize$\pm$0.36} & 58.91\,{\scriptsize$\pm$0.36} & 41.97\,{\scriptsize$\pm$0.49} & 59.93\,{\scriptsize$\pm$0.69} & 63.43\,{\scriptsize$\pm$0.21} & 51.81\,{\scriptsize$\pm$0.39} \\
    9 local crops & 64.66\,{\scriptsize$\pm$0.31} & 71.08\,{\scriptsize$\pm$0.23} & 51.90\,{\scriptsize$\pm$0.41} & 72.07\,{\scriptsize$\pm$1.55} & 74.08\,{\scriptsize$\pm$0.25} & 61.79\,{\scriptsize$\pm$0.78} \\
    \bottomrule
    \end{tabular}
\end{table}

\subsection{Cross-Tokenizer Token Localization on COCO}\label{app:coco_localization}

CSSD contains single salient objects and the analysis above compares \methodName variants only. To test whether individual tokens localize objects in multi-object images, and how this compares to prior tokenizers, we evaluate on a fixed random subset of 150 COCO~2017 validation images~\citep{lin2014microsoft}. We compare only tokenizers that produce 256 tokens per image: since the score below takes the maximum over individual tokens, it grows with the number of candidate tokens and would favor tokenizers with longer sequences. We restrict the evaluation to large objects, i.e., instances whose original COCO annotation area is at least $96^2$ pixels; crowd annotations are excluded, and an instance is retained only if its mask remains non-empty after square center-cropping and resizing the image to $256 \times 256$. This yields 230 objects in 119 images. The restriction follows from how a single token map is scored. A map that retains the top $Q\%$ of the pixels always covers $Q\%$ of the image, so its IoU with an object that covers a fraction $f$ of the image is at most $\min(Q, f) / \max(Q, f)$, and an IoU above 0.5 is only possible if $Q/2 < f < 2Q$. Medium objects ($32^2 \leq \text{area} < 96^2$) cover a median of 1.33\% of the resized image, and 90\% of them less than 3.54\%, so only maps that keep a few cells of the $16 \times 16$ attention grid could localize them. Accordingly, every evaluated model localizes fewer than 10\% of the medium objects at the IoU threshold of 0.5, and this setting does not discriminate between tokenizers. Large objects cover a median of 12.73\% of the image.

\noindent\textbf{Protocol.} Every tokenizer completes its full 20-step decoding trajectory from its token sequence, and attention maps are captured at the same clean-progress value of 0.1 for all models (native $t = 0.1$ for \methodName and native noise time $0.9$ for the rectified-flow decoder of FlexTok) rather than at unmatched native timesteps; the decoding noise is fixed per image. As for CSSD, the maps are the raw decoder spatial-query/token-key attention logits on the $16 \times 16$ spatial grid, softmaxed over spatial positions separately for every token, averaged over heads and bilinearly resized to $256 \times 256$. A token map is binarized by retaining its top $Q\%$ of pixels. For each ground-truth object $o$ with mask $S_o$ we compute
\begin{align}
    {\rm IoU}_o = \max_i \, {\rm IoU}(M_i, S_o),
\end{align}
where $M_i$ is the binary attention map of the individual token $i$, and count the object as localized when ${\rm IoU}_o > p$. This evaluation uses no learned readout and never combines several token maps. Based on the size analysis above, we fix $Q = 10\%$ for all tokenizers: a map that retains 10\% of the pixels can exceed an IoU of 0.5 for objects that cover between 5\% and 20\% of the image, a range that contains the median large object, and a common $Q$ makes every tokenizer target the same object sizes. The extraction layer is calibrated independently for each tokenizer, on a disjoint set of 13 images containing 25 large objects, by sweeping every decoder layer and maximizing the localization rate at the primary threshold $p = 0.5$ (ties are broken by the mean best-token IoU, then by earlier layers). This selected the zero-based layers 1, 4, 4, 4, 12 and 15 for \methodName-B, its variants without SREPA and without local crops, \methodName-L, \methodName-XL and FlexTok, respectively.

\begin{table}[h]
    \centering
    \footnotesize
    \setlength{\tabcolsep}{3pt}
    \caption{Individual-token localization of the 230 large objects in the COCO subset. An object is localized if the best of the individual binarized token attention maps has ${\rm IoU} > p$ with its mask. Intervals are 95\% bootstrap intervals of the rate at $p = 0.5$. The last two columns count how many distinct token indices supply the best match over all 230 objects, and the normalized entropy of this distribution. All tokenizers produce 256 tokens per image.}
    \label{tab:coco_localization}
    \begin{tabular}{@{}lccccccc@{}}
    \toprule
    & \multicolumn{2}{c}{Localized at $p = 0.5$} & Mean best & \multicolumn{2}{c}{Localized at} & Distinct & Norm. \\
    Tokenizer & rate$\uparrow$ & 95\% interval & IoU$\uparrow$ & $p = 0.25$ & $p = 0.75$ & best tokens & entropy \\
    \midrule
    FlexTok{\tiny (d18-d18)} & 21.74 & $[16.39, 27.44]$ & 0.386 & 79.13 & \second{2.17} & 59 / 256 & 0.662 \\
    \midrule
    \methodName-B, no local crops & 17.39 & $[12.86, 22.22]$ & 0.358 & 74.78 & 1.30 & 127 / 256 & 0.843 \\
    \methodName-B, no SREPA & 23.48 & $[18.14, 29.17]$ & 0.386 & 80.87 & 0.87 & 117 / 256 & 0.818 \\
    \methodName-B (ours) & 25.65 & $[19.72, 31.69]$ & 0.406 & \second{83.91} & 0.43 & 134 / 256 & 0.854 \\
    \methodName-L (ours) & \best{36.52} & $[30.13, 43.22]$ & \best{0.433} & \best{84.35} & \best{3.48} & 126 / 256 & 0.836 \\
    \methodName-XL (ours) & \second{34.78} & $[28.71, 41.07]$ & \second{0.423} & \second{83.91} & 1.30 & 125 / 256 & 0.836 \\
    \bottomrule
    \end{tabular}
\end{table}

\noindent\textbf{Results.} Table~\ref{tab:coco_localization} summarizes the evaluation. \methodName-L localizes 14.78 points more large objects than FlexTok-d18-d18~\citep{bachmann2025flextok} (paired bootstrap 95\% interval $[+9.33, +20.38]$), has a higher mean best-token IoU (0.433 versus 0.386), and remains ahead at all three thresholds. \methodName also distributes the best matches over many more token indices than FlexTok (126 versus 59 of 256 tokens; normalized entropy 0.836 versus 0.662), i.e., localization is not carried by a few dedicated slots. The distribution statistics alone do not establish semantic disentanglement, but together with the higher per-object localization rates they support the intended effect of recurrent aggregation: spatial information is organized into more broadly utilized, object-aligned tokens. Within the B scale, recurrent local-crop training raises the localization rate from 17.39\% to 25.65\% (paired interval $[+1.42, +14.83]$) and the mean best-token IoU from 0.358 to 0.406 under the common mask size, so that \methodName-B numerically surpasses the larger FlexTok-d18-d18 (25.65\% versus 21.74\%), whereas its no-local variant does not. The no-SREPA variant of \methodName-B (Table~\ref{tab:srepa_ablation}) localizes 23.48\% of the objects with a mean best-token IoU of 0.386. Its paired difference to the full model ($-2.17$ points, interval $[-8.30, +3.78]$; mean best-token IoU $-0.020$, interval $[-0.032, -0.008]$) is much smaller than that of the no-local variant, which it exceeds by 6.09 points (interval $[+0.47, +11.62]$). This variant is also the one whose semantic supervision matches that of FlexTok: both are trained with standard REPA on DINOv2 features and without aggregate message-level alignment. Under this matched supervision, \methodName-B without SREPA localizes large objects as often as FlexTok-d18-d18 (23.48\% versus 21.74\%; paired difference $+1.74$ points, interval $[-3.90, +7.23]$) with the same mean best-token IoU (0.386), although it has about a fifth of FlexTok's parameters (174M versus 860M, Table~\ref{tab:main_comparisons}), and it spreads the best matches over twice as many token indices (117 versus 59 of 256; normalized entropy 0.818 versus 0.662). Token-level localization in \methodName thus does not rely on SREPA. The multi-object diagnostic thus reproduces the CSSD results of Section~\ref{sec:ablations}: removing SREPA barely changes token--object localization, whereas removing recurrent local crops degrades it substantially.

Taken together, the probes of this paper address distinct levels of the claim: the shared sequence probes of Table~\ref{tab:main_comparisons} and Appendix~\ref{app:affine_probe} measure the accessibility of semantic information from the complete message, the segmentation probe below measures the dense spatial information available to a common decoder, and CSSD together with this analysis measure the alignment of individual tokens with objects.

\subsection{Segmentation Probe}\label{app:segmentation_probe}

The segmentation probe of Section~\ref{sec:quantitative} evaluates \methodName-L and SelfTok-512 on the pair-disjoint MSCOCO subset of Appendix~\ref{app:test_suite}. The target for each image is a binary foreground mask formed as the union of all COCO instance masks; images and masks are resized and center-cropped to $256 \times 256$, and the targets are downsampled to the output resolution of the probe. \methodName-L encodes only the global crop, without test-time aggregation over local crops. Both tokenizers are frozen and read out by the same lightweight decoder: the tokens are normalized and linearly projected to 128 dimensions, 256 learned spatial queries attend to them through two cross-attention blocks with four heads each, and the queries are arranged into a $16 \times 16$ feature map that is decoded to mask logits by an upsampling convolutional stem. Both probes use the same optimization protocol: AdamW for at most 20 epochs, batch size 256, learning rate 1e-3, weight decay 1e-4, early stopping with a patience of five epochs, and a binary cross-entropy plus Dice loss. At a prediction threshold of 0.5, \methodName-L obtains a foreground IoU of 0.532, compared with 0.436 for SelfTok-512, despite SelfTok's much stronger reconstruction. Because the spatial queries can attend across all tokens and the target merges instances, this probe measures the spatial information retained by the complete token sequence; it is complementary to the individual-token analyses of Appendices~\ref{app:cssd_protocol} and~\ref{app:coco_localization} and is not evidence for a one-token-one-instance decomposition. However, because the probe is trained and evaluated on the pair-disjoint MSCOCO split, it also tests compositional generalization.

%% file: sections_arxiv/appendix/extended_related_work.tex
\section{Extended Related Work}\label{app:extended_related_work}

Here we expand the related work of Section~\ref{sec:related_work} with recent and concurrent tokenizers. GigaTok~\citep{xiong2025gigatok} scales 1D tokenizers to billions of parameters and regularizes their latents with DINOv2 features to resolve the reconstruction--generation dilemma of large tokenizers; its contribution is scale rather than the organization of the token sequence. VFMTok~\citep{zheng2025vfmtok} starts from the frozen spatial features of a vision foundation model and adaptively resamples them into tokens, whereas \methodName learns its discrete message end-to-end from pixels. SemTok~\citep{qu2026semtok} learns 1D tokens with a single-pass encoder, SigLIP-based semantic alignment and a large pretrained decoder that is trained in two stages, namely generation pretraining followed by reconstruction refinement. CaTok~\citep{chen2026catok} learns causal register tokens whose intervals are tied to MeanFlow time intervals from a single global pass over the image, and its main representation is continuous. EvoTok~\citep{li2026evotok} evolves residual latents over a 2D token grid, with deeper residual levels becoming more semantic, and targets a unified understanding-and-generation model trained on much larger data. VibeToken~\citep{patel2026vibetoken} scales 1D tokenizers and autoregressive models to dynamic resolutions and aspect ratios without a semantic teacher. PyraTok~\citep{susladkar2026pyratok} and PVQ-VAE~\citep{prakash2026pyramidal} are video tokenizers built on pyramidal representations.

These works span 1D and 2D codes, discrete and continuous representations, image and video modalities, and ImageNet-scale versus web-scale training. What distinguishes \methodName is the mechanism it studies: recurrent partial observations that repeatedly rewrite a fixed-length discrete message, together with the frozen-token evaluation of the resulting token organization.

%% file: sections_arxiv/appendix/more_visual_results.tex
\section{More Visual Results}\label{app:visual_results}

Here we provide more qualitative results with \methodName that could not be included in the main paper due to the pages number limit. If not stated otherwise, the input images in the main paper and in this appendix are random images from \url{unsplash.com}. We observed that despite being trained only on ImageNet1k~\cite{deng2009imagenet}, \methodName demonstrates remarkable generalization capabilities and remains robust under domain shifts (see Figure~\ref{fig:ood}).

\begin{figure}
    \centering
    \newcommand{\curWidth}{0.17\linewidth}
    \begin{tabular}{@{}c|cccc@{}}
         Query image & 1st neighbor & 2nd neighbor & 3rd neighbor & 4th neighbor \\
         \midrule
         \includegraphics[width=\curWidth]{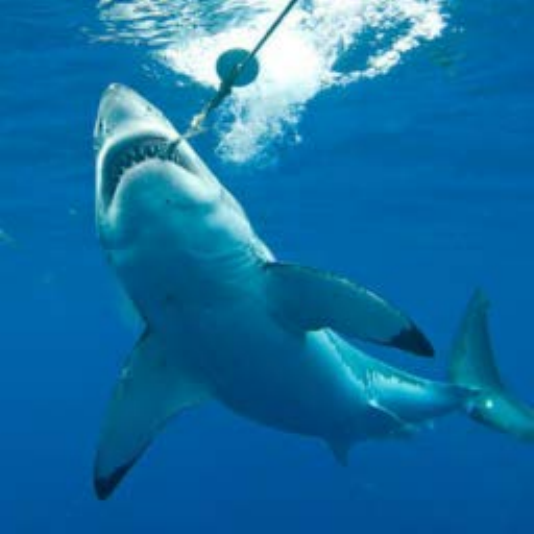} &
         \includegraphics[width=\curWidth]{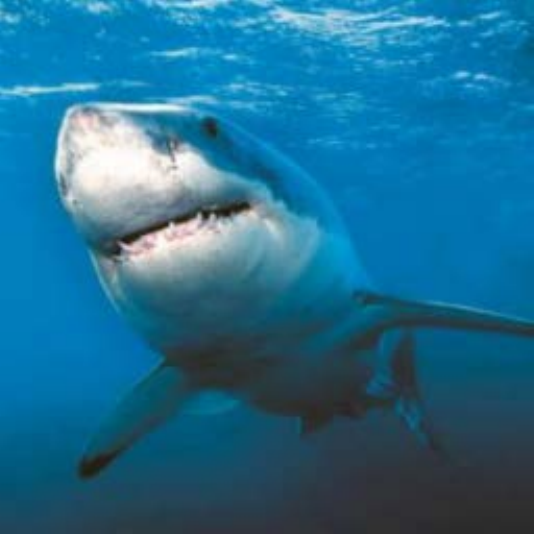} &
         \includegraphics[width=\curWidth]{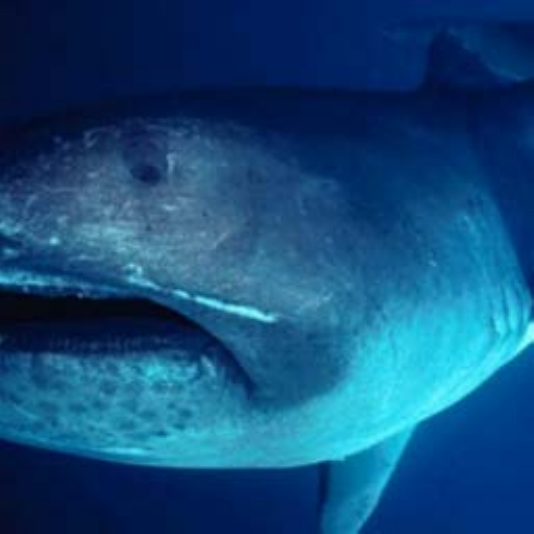} &
         \includegraphics[width=\curWidth]{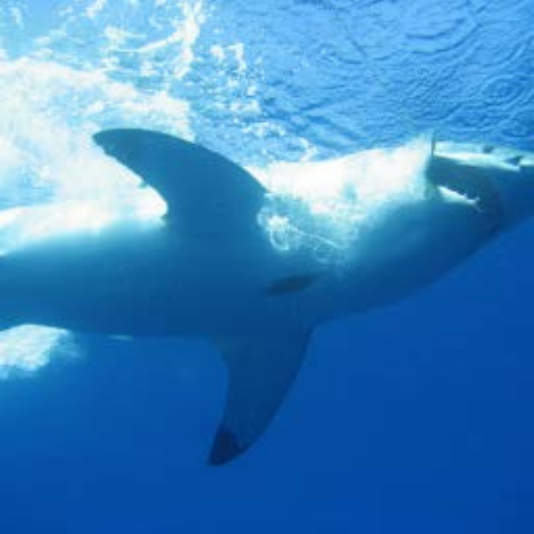} &
         \includegraphics[width=\curWidth]{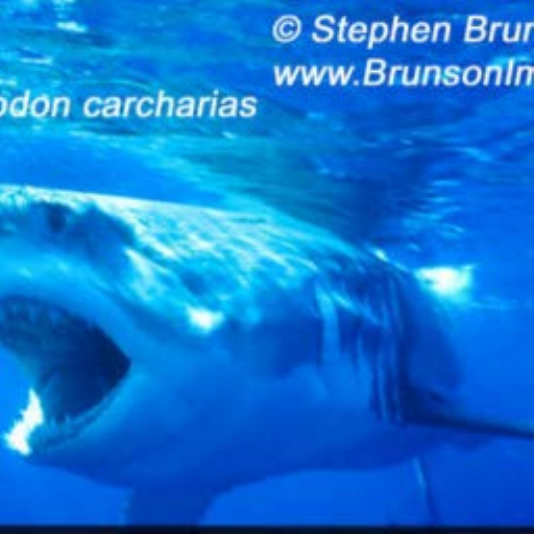} \\
         
         \includegraphics[width=\curWidth]{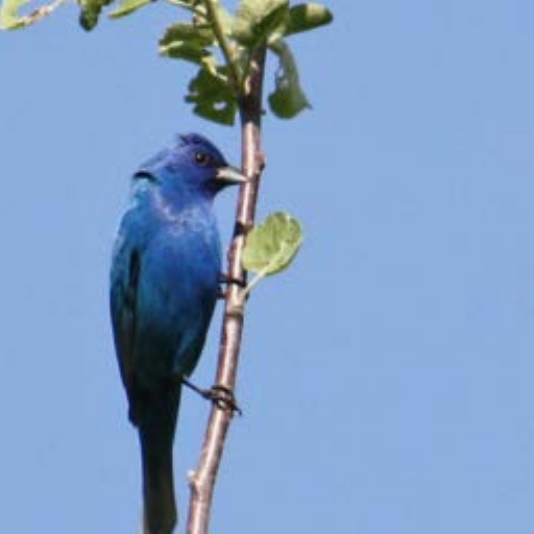} &
         \includegraphics[width=\curWidth]{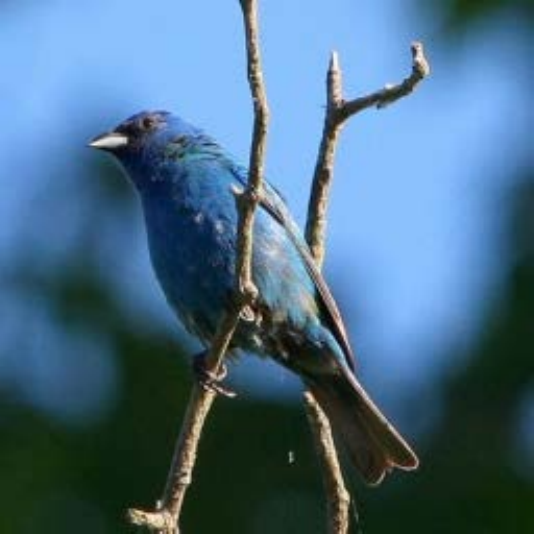} &
         \includegraphics[width=\curWidth]{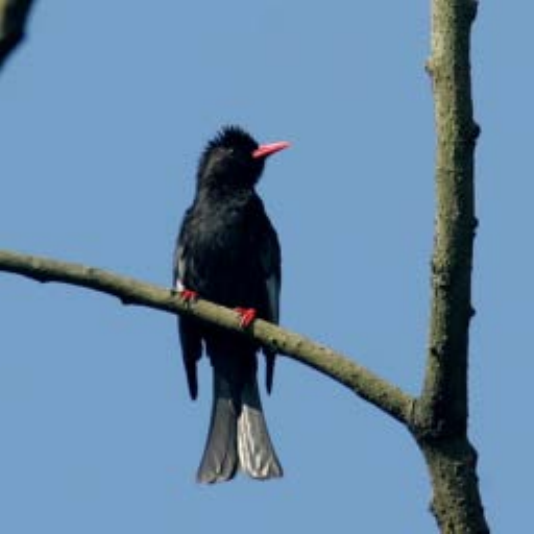} &
         \includegraphics[width=\curWidth]{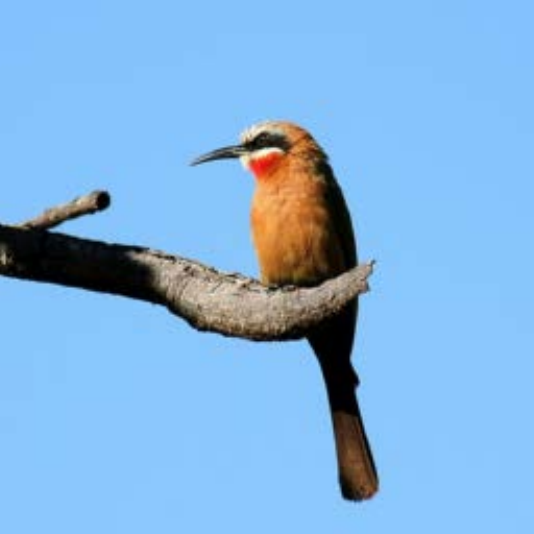} &
         \includegraphics[width=\curWidth]{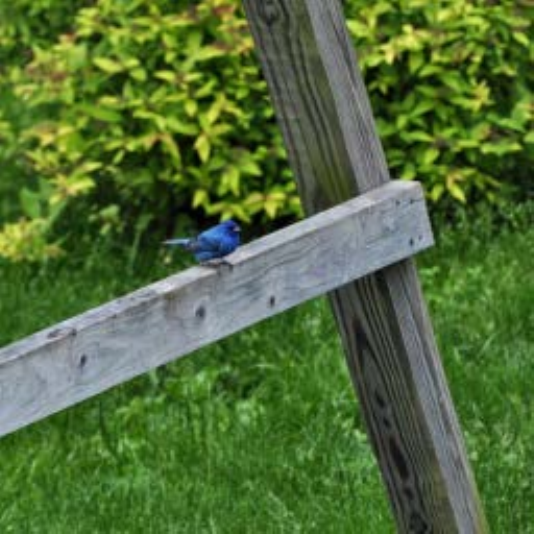} \\

         \includegraphics[width=\curWidth]{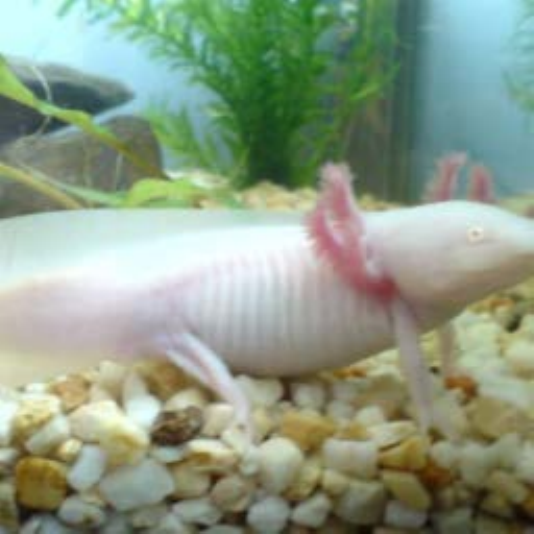} &
         \includegraphics[width=\curWidth]{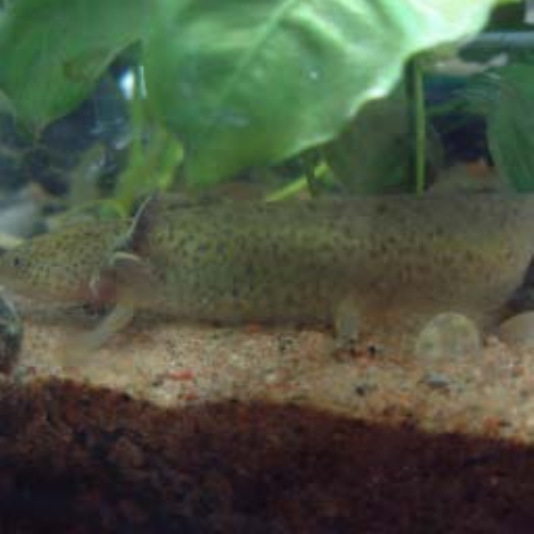} &
         \includegraphics[width=\curWidth]{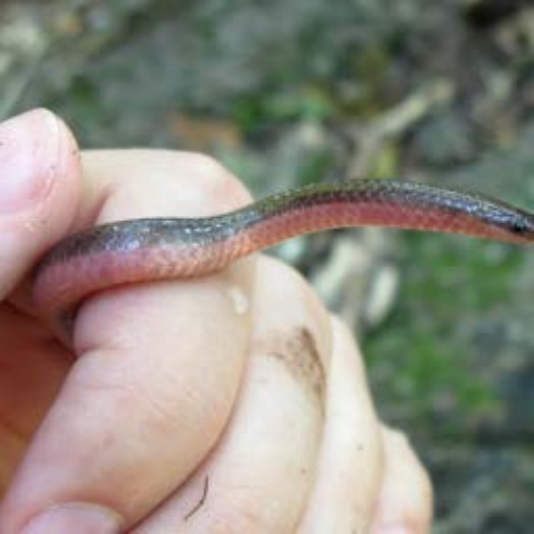} &
         \includegraphics[width=\curWidth]{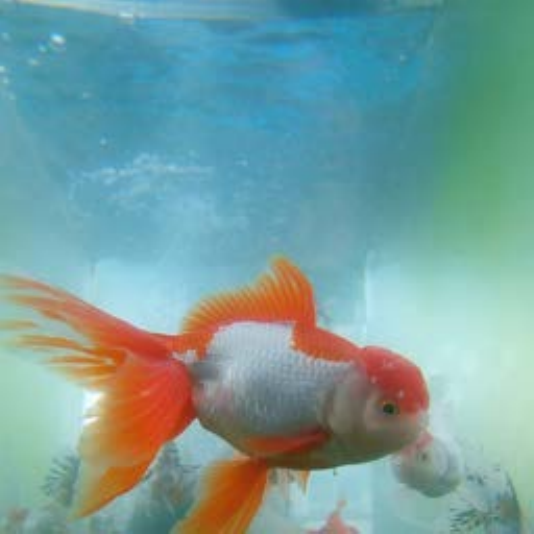} &
         \includegraphics[width=\curWidth]{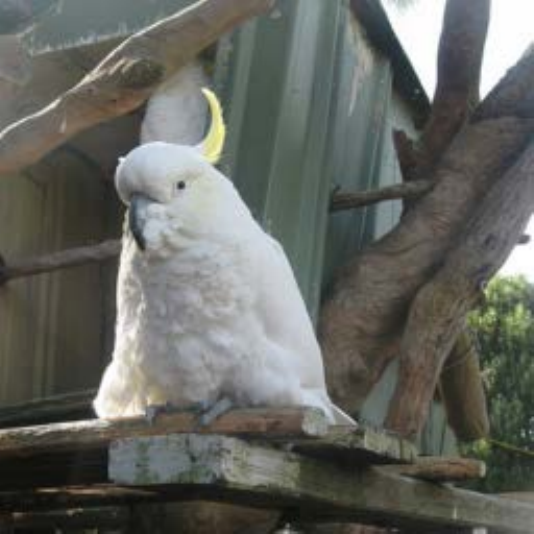} \\

         \includegraphics[width=\curWidth]{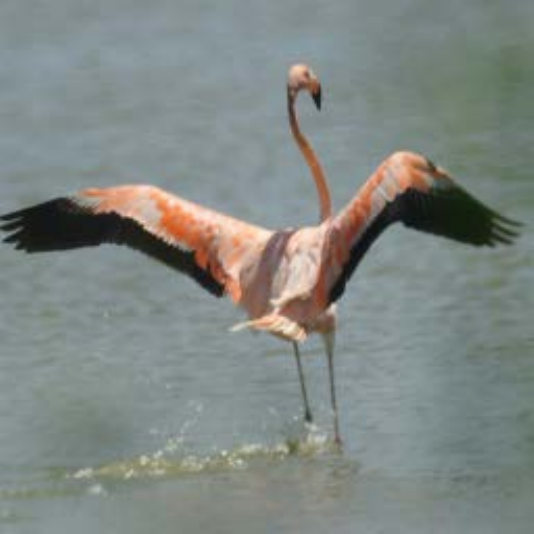} &
         \includegraphics[width=\curWidth]{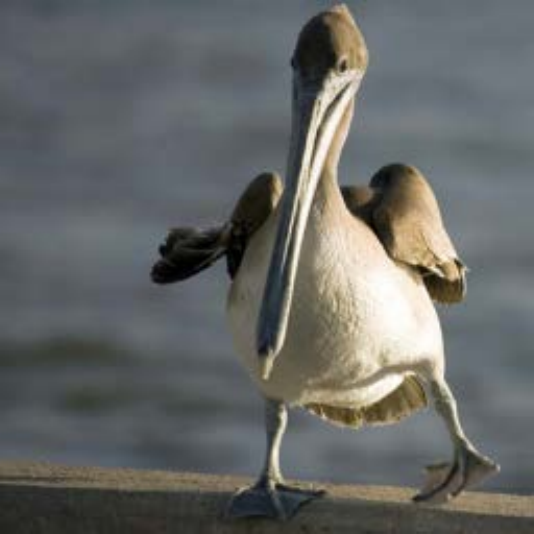} &
         \includegraphics[width=\curWidth]{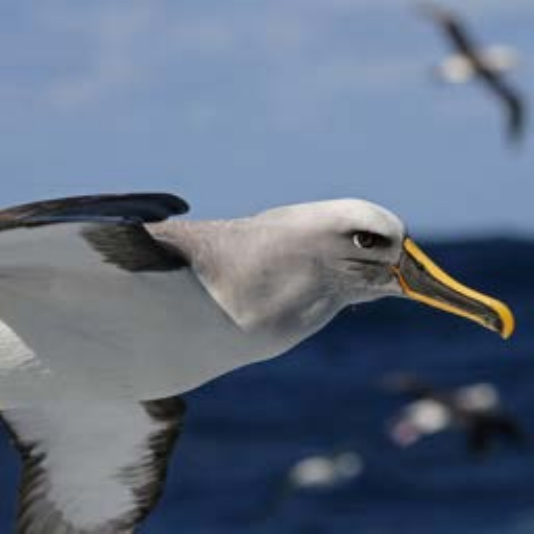} &
         \includegraphics[width=\curWidth]{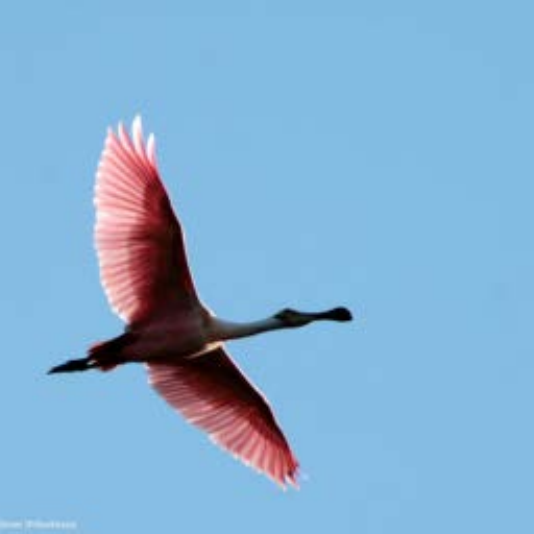} &
         \includegraphics[width=\curWidth]{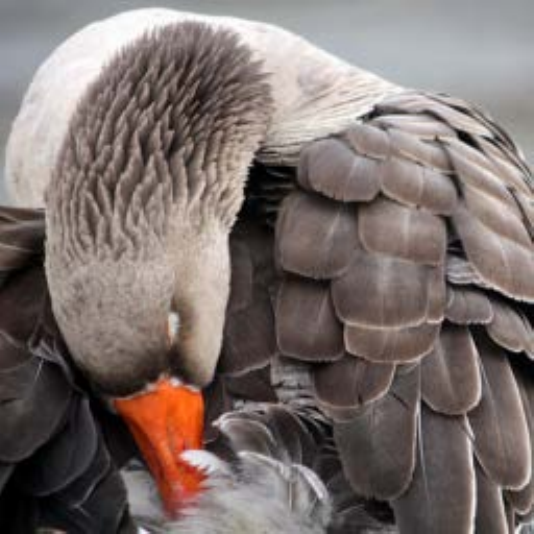} \\
         
         \includegraphics[width=\curWidth]{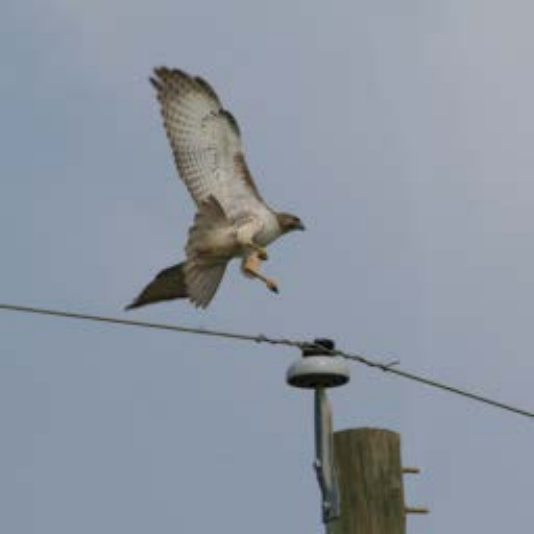} &
         \includegraphics[width=\curWidth]{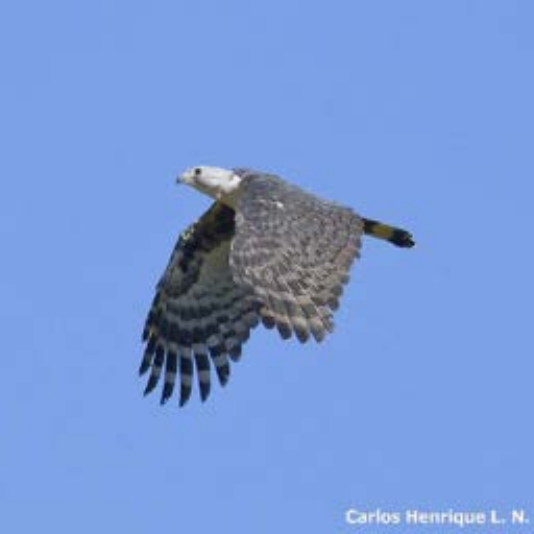} &
         \includegraphics[width=\curWidth]{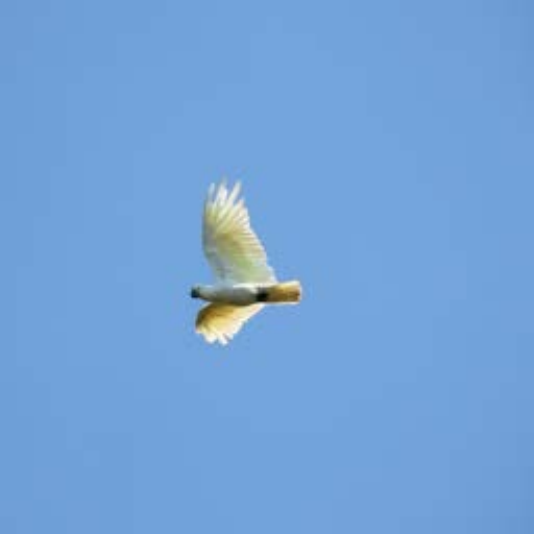} &
         \includegraphics[width=\curWidth]{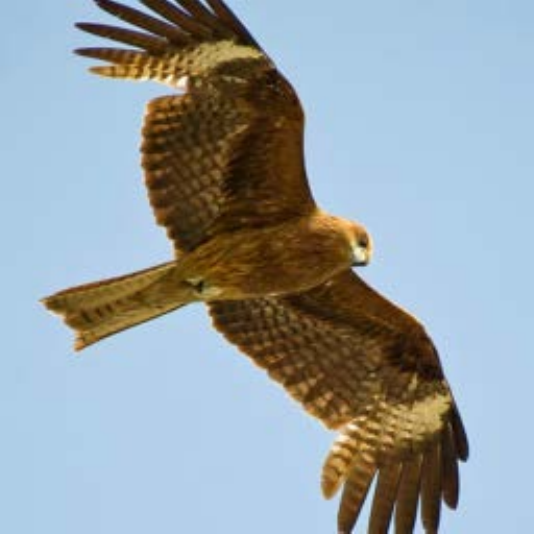} &
         \includegraphics[width=\curWidth]{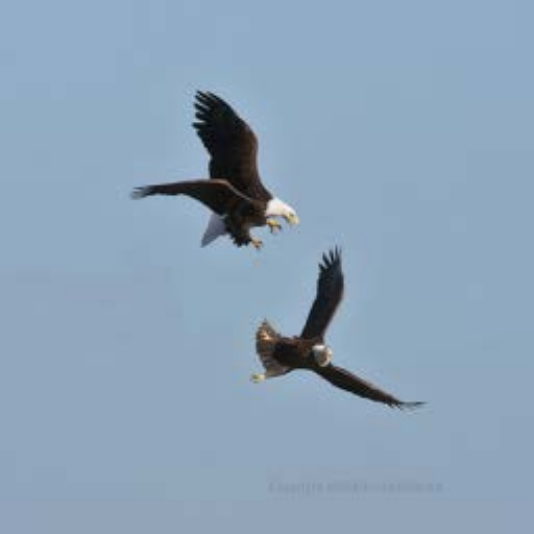} \\
         
         \includegraphics[width=\curWidth]{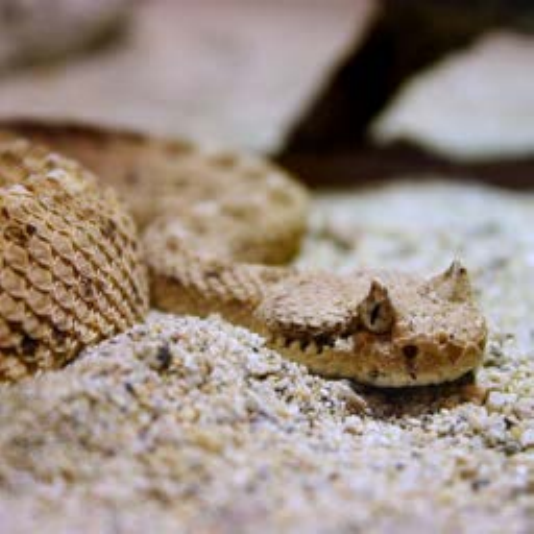} &
         \includegraphics[width=\curWidth]{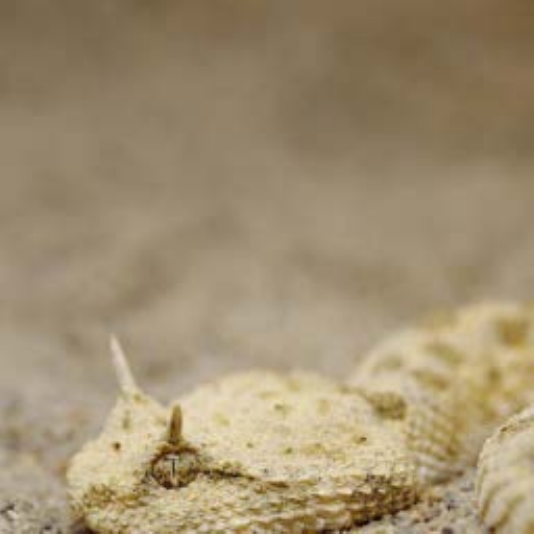} &
         \includegraphics[width=\curWidth]{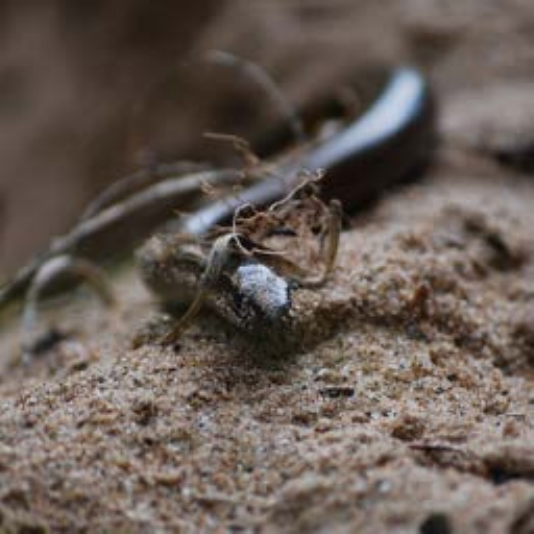} &
         \includegraphics[width=\curWidth]{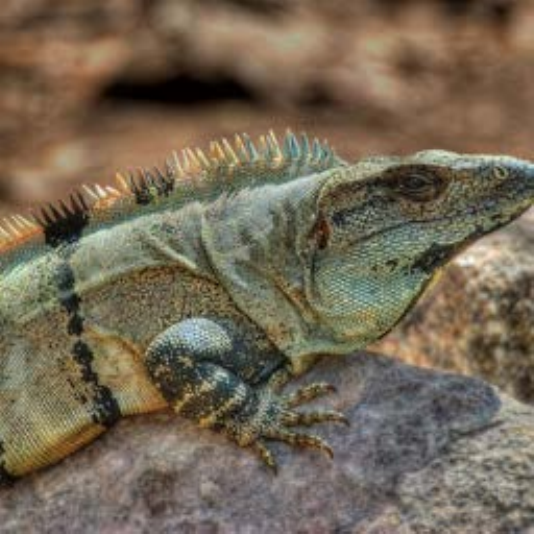} &
         \includegraphics[width=\curWidth]{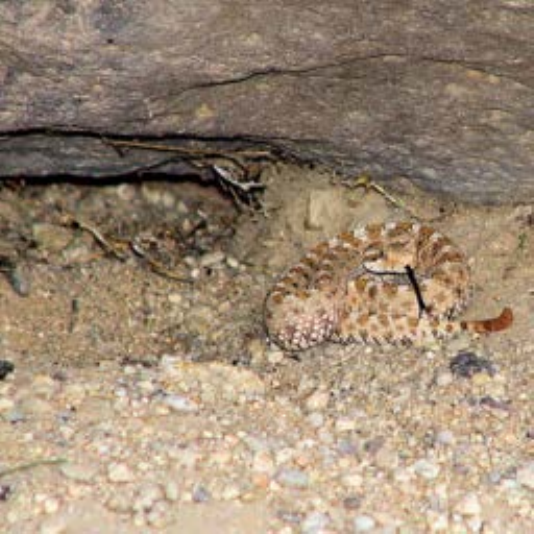} \\
         
         \includegraphics[width=\curWidth]{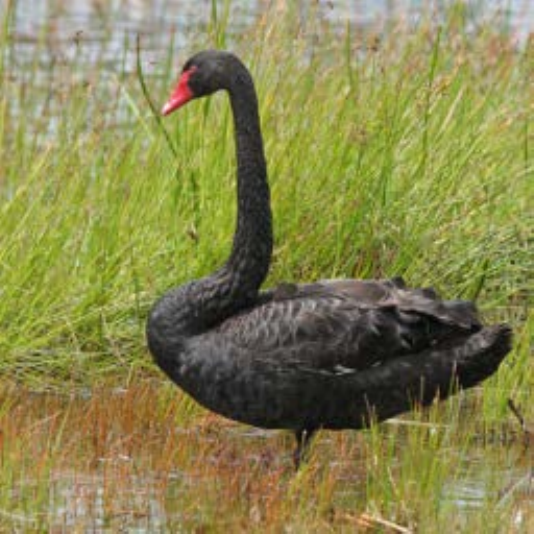} &
         \includegraphics[width=\curWidth]{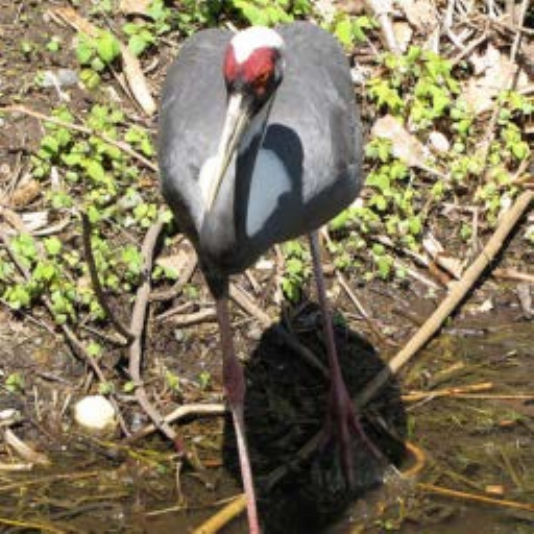} &
         \includegraphics[width=\curWidth]{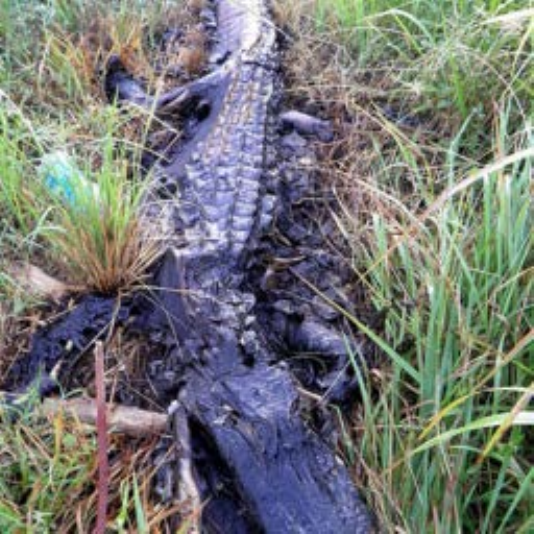} &
         \includegraphics[width=\curWidth]{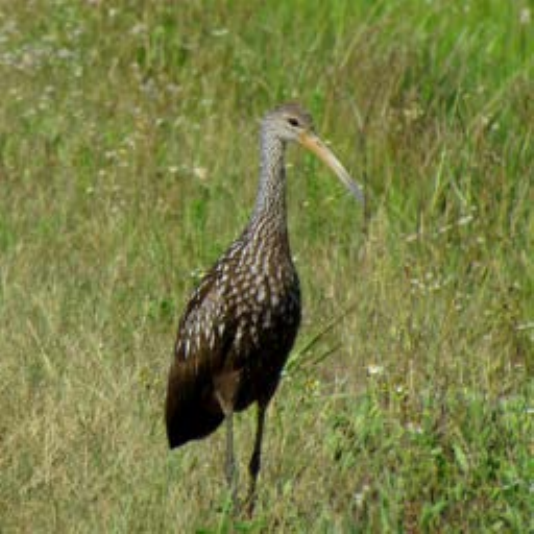} &
         \includegraphics[width=\curWidth]{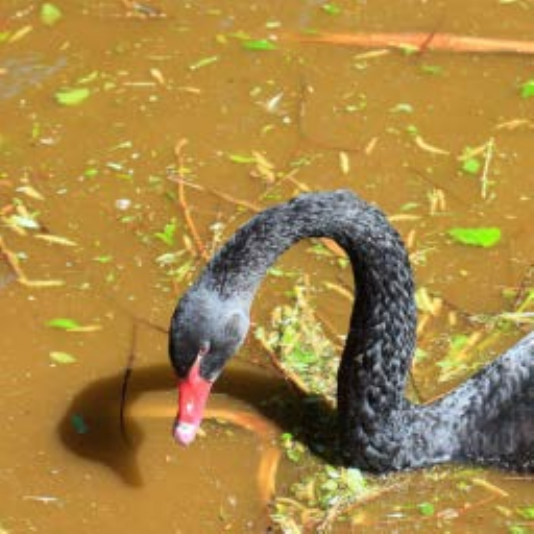} \\
    \end{tabular}
    \caption{Visualization of the nearest neighbors of a sample from the ImageNet100 validation set in the space of \methodName's latent messages. One can see that despite unsupervised training and simple probing the learned messages tend to cluster into semantically meaningful groups.}
    \label{fig:neighbors}
\end{figure}

\noindent\textbf{Nearest Neighbors.} To study the learned space of latent messages, we visualize some nearest neighbors of sample images from the validation set of ImageNet100~\citep{deng2009imagenet}. For simplicity, we concatenate the tokens in latent messages to obtain a single 1536-dimensional vector per image. For each query image we then find 4 closest samples among the whole validation set using the cosine distance as our similarity measure (see Figure~\ref{fig:neighbors}). Despite the simplicity of this probing mechanism, we find that the latent space is well--structured and the neighboring images typically share semantics.

\begin{figure}[t]
    \centering
    \includegraphics[width=0.98\linewidth, clip, trim=0cm 0.9cm 0cm 0cm]{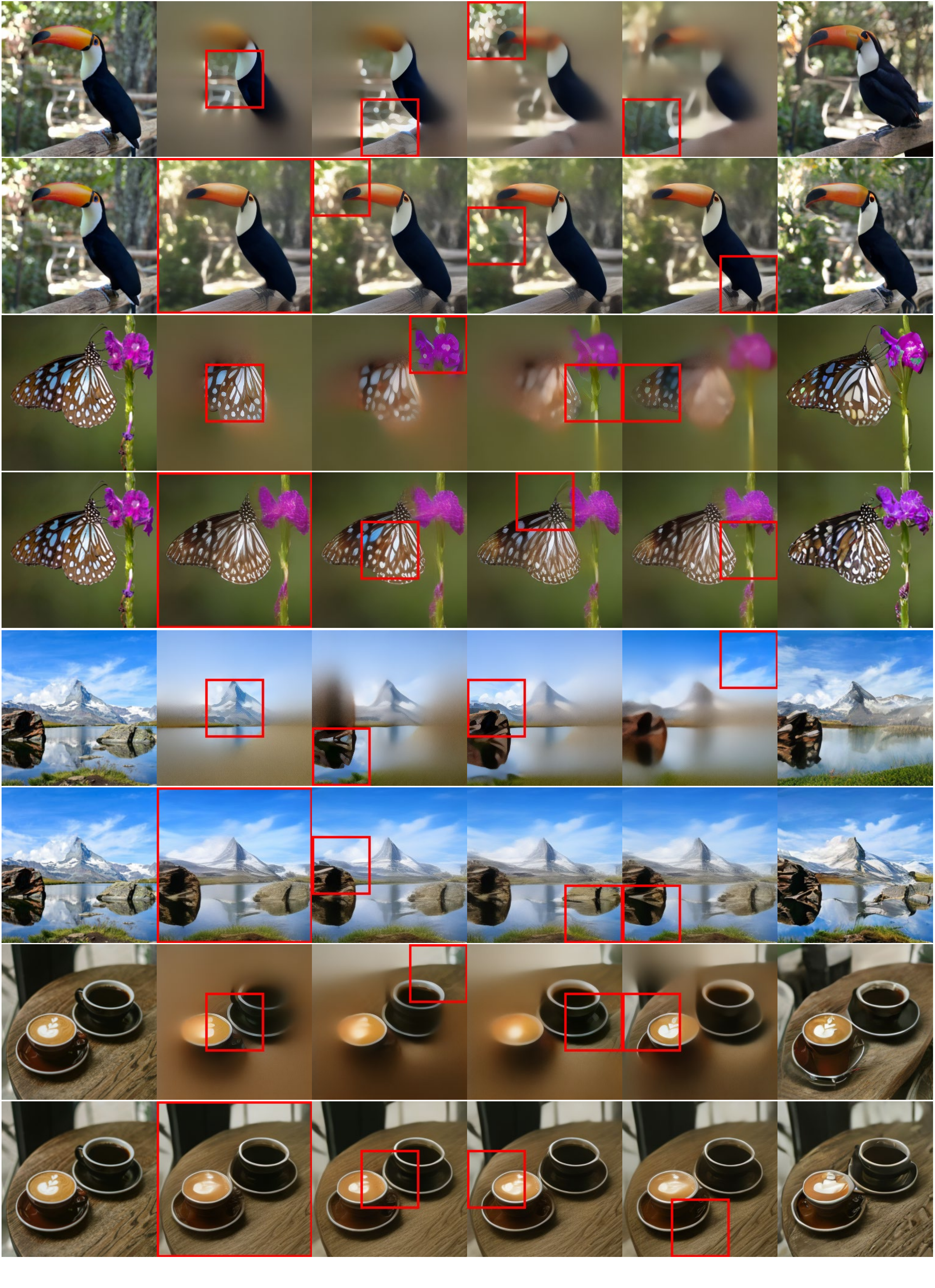}
    \caption{Additional results on the difference between the \textit{adaptive} (odd rows) and \textit{global+adaptive} (even rows) cropping policies. The first column depicts the ground truth input image. The last column corresponds to the final reconstruction with 10 NFE. The columns in-between demonstrate how \methodName refines its latent message with incoming information (1 NFE reconstructions are shown).}
    \label{fig:more_global_local}
\end{figure}

\noindent\textbf{Cropping policies.} Figure~\ref{fig:global_vs_local} demonstrates the difference between adaptive cropping policies in the absence and the presence of the global crop. When the global crop is leveraged, it provides a way to quickly embed the whole scene into the latent message with the subsequent local crops allowing to focus on the finer details (such as the shape and the color of the feathers near the bird's head). Figure~\ref{fig:more_global_local} contains more demonstrations on the difference between the presence and absence of the global crop in the adaptive cropping policy. Notice that the content that the model does not observe in the sequence of crops misses from the final reconstruction. However, 3-4 local crops in the adaptive policy are typically enough to grasp the overall idea behind the scene without including too much details. This explains, why the adaptive policy with only 3 crops performs quite well on our semantic probing test suite.


%% file: sections_arxiv/appendix/future_work.tex
\section{Future Work}\label{app:future_work}

Going forward, an important direction is to improve reconstruction fidelity without sacrificing semantic structure, making \methodName more suitable for generation-focused applications. More broadly, structured token sequences could provide a stronger interface for multimodal models, especially in settings that require object-centric reasoning, compositional understanding, and dense visual grounding. The test-time flexibility of \methodName also suggests opportunities for adaptive tokenization, where cropping policies are selected dynamically for the task.

Future work will extend \methodName to video, where temporal redundancy makes compact and structured representations especially valuable. By accumulating spatiotemporal observations over time, \methodName could learn token sequences that capture motion, actions, and long-range temporal structure, enabling more efficient and semantically grounded video understanding.

%% file: sections_arxiv/appendix/impact_statement.tex
\section{Impact Statement}\label{app:impact}

This paper advances research in autoencoders, representation learning and generative modeling. As with other generative image models and tokenizers, the methods studied here may carry standard societal and ethical risks. These include the potential misuse of generated content for creating misleading or deceptive media, the amplification of biases present in training data, and the environmental costs associated with training and deploying large-scale models. While this work is primarily methodological, we acknowledge these well-documented considerations and emphasize that responsible dataset curation, evaluation, and deployment practices are important to mitigate such risks.